\documentclass{article}

\usepackage{microtype}
\usepackage{graphicx}
\usepackage{subcaption}
\usepackage{booktabs} 

\usepackage{hyperref}

\usepackage[accepted]{icml2026}

\usepackage{amsmath}
\usepackage{amssymb}
\usepackage{mathtools}
\usepackage{amsthm}

\usepackage{algorithm}
\usepackage[noend]{algpseudocode}  
\usepackage{amsmath}
\usepackage[capitalize,noabbrev]{cleveref}

\theoremstyle{plain}

\theoremstyle{definition}

\theoremstyle{remark}

\usepackage[textsize=tiny]{todonotes}

\icmltitlerunning{Stabilizing Native Low-Rank LLM Pretraining}

\newcommand{\norm}[1]{\left\lVert #1 \right\rVert}
\newcommand{\abs}[1]{\left| #1 \right|}
\begin{document}

\twocolumn[
  \icmltitle{Stabilizing Native Low-Rank LLM Pretraining}



  \icmlsetsymbol{equal}{*}

  \begin{icmlauthorlist}
    \icmlauthor{Paul Janson}{con,mila}
    \icmlauthor{Edouard Oyallon}{cnrs}
    \icmlauthor{Eugene Belilovsky}{con,mila}
  \end{icmlauthorlist}

  \icmlaffiliation{con}{Concordia University, Montreal, Canada}
  \icmlaffiliation{mila}{Mila Quebec AI Institute, Montreal, Canada}
  \icmlaffiliation{cnrs}{Sorbonne University, CNRS , Paris , France}

  \icmlcorrespondingauthor{Paul Janson}{paul.janson@mila.quebec}

  \icmlkeywords{Machine Learning, ICML}

  \vskip 0.3in
]



\printAffiliationsAndNotice{}  

\begin{abstract}

Foundation models have achieved remarkable success, yet their growing parameter counts pose significant computational and memory challenges. Low-rank factorization offers a promising route to reduce training and inference costs, but the community lacks a stable recipe for training models from scratch using exclusively low-rank weights while matching performance of the dense model. 
We demonstrate that Large Language Models (LLMs) can be trained from scratch using exclusively low-rank factorized weights for all non-embedding matrices without auxiliary ``full-rank" guidance required by prior methods. While native low-rank training often suffers from instability and loss spikes, we identify uncontrolled growth in the spectral norm (largest singular value) of the weight matrix update as the dominant factor. To address this, we introduce \textbf{Spectron:} \textbf{Spectr}al renormalization with orthogonalizati\textbf{on}, which dynamically bounds the resultant weight updates based on the current spectral norms of the factors. Our method enables stable, end-to-end factorized training with negligible overhead. Finally, we establish compute-optimal scaling laws for natively low-rank transformers, demonstrating predictable power-law behavior and improved inference efficiency relative to dense models.
Our code is available at \href{https://github.com/Pauljanson002/spectron}{https://github.com/Pauljanson002/spectron}

\end{abstract}

\section{Introduction}
\label{sec:intro}

\begin{figure}
    \centering
    \includegraphics[width=0.9\linewidth]{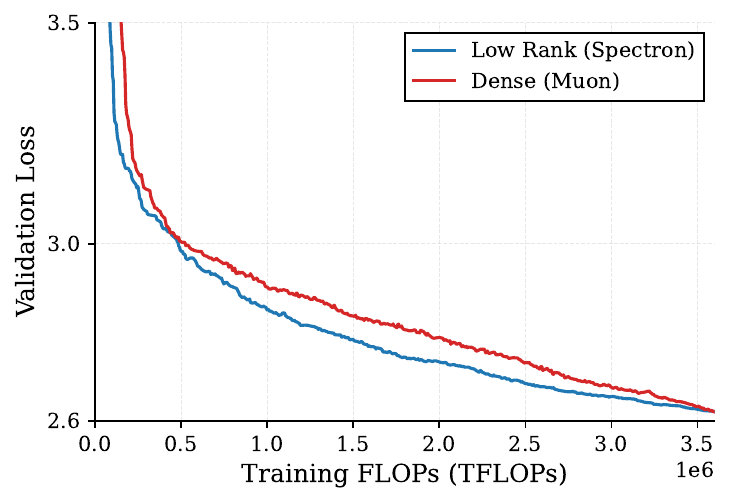}
    \caption{\textbf{Natively Low-Rank Training Achieves Dense-Level Performance.} Validation loss curves comparing a 780M dense Transformer~\cite{vaswani2017attention}(\textcolor{red}{red}) against our 454M low-rank factorized Transformer (\textcolor{blue}{blue}) across $3.5 \times10^6$ training TFLOPs on FineWeb~\cite{penedo2024fineweb}. Our method \textbf{Spectron} enables stable end-to-end factorized training that matches dense performance at equal compute, yielding an inference-optimal model with substantially fewer parameters.}
    \label{fig:flops_teaser}
    \vspace{-8mm}
\end{figure}

Foundation models continue to scale in size, with model capacity remaining a primary driver of performance improvements in frontier systems~\citep{kaplan2020scaling,hoffmann2022an,wei2022emergent,brown2020language}. 
Training these dense architectures, however, presents substantial computational bottlenecks, particularly regarding  memory constraints that motivate parallelization~\cite{huh2024training,shazeer2017,rajbhandari2020zero,rivaudpetra,nabli2025acco} and factorization~\cite{khodakinitialization,wei2024building,chollet2017xception} strategies that distribute computation across accelerators while preserving throughput. Low-rank parameterization, where we express weight matrices as $W=AB^\top$ with $A \in \mathbb{R}^{m \times r}$ and $B \in \mathbb{R}^{n \times r}$ for $r < min(m,n)$  has achieved significant success in fine-tuning~\cite{hu2022lora,dettmers2023qlora} by reducing both memory footprint and floating-point operations required for adapter-style updates. Yet practitioners have rarely applied this approach to pretraining large language models (LLMs), primarily because existing methods~\citep{wei2024building,wang2021pufferfish,huh2024training} rely on workarounds that keep full-rank auxiliary weights, causing minimal adoption. A stable recipe for training factorized layers in modern transformer architectures~\citep{touvron2023llama,vaswani2017attention} has remained elusive.

Empirical evidence demonstrates that models tend to converge toward low-rank representations by the end of training~\citep{martin2021implicit,pmlr-v280-galanti25a,yang2023spectral,ramasinghe2025subspace}. Neural networks exhibit high compressibility~\citep{lecun1989optimal,franklelottery,yu2017compressing} and respond effectively to low-rank fine-tuning updates~\citep{hu2022lora,dettmers2023qlora}, suggesting that full-rank parameterization maintained throughout pretraining can be traded for low-rank representations to reduce the memory requirements. This observation motivates a fundamental question: \emph{Can we train foundation models directly in a low-rank factorized regime from initialization while maintaining competitive performance?}

Answering this question carries both scientific and practical significance.
Scientifically, exploiting low-rank structure during pretraining would provide insights into the learning dynamics of modern transformer architectures~\cite{touvron2023llama,vaswani2017attention} and reveal whether full-rank representations are necessary during optimization.
Practically, native low-rank training could democratize foundation model development by reducing hardware requirements, potentially decreasing memory consumption by factors proportional to the rank reduction. 

 Training neural networks with native low-rank parameterization from scratch confronts severe instabilities when we factorize all non-embedding weight matrices to low rank. Unlike LoRA~\citep{hu2022lora}, which preserves frozen full-rank model weights, this approach encounters fundamental difficulties arising from how low-rank parameterization interacts with optimization dynamics. 
 When we update factors independently, a given factorization  $W=AB^\top$ permits infinitely many equivalent representations $W=(\lambda.A)(\frac 1\lambda .B)$ for any $\lambda > 0$. This scaling invariance permits unbounded growth in $\lambda$
, leading to unrestricted spectral norm (largest singular value) expansion, triggering exploding activations, and ultimately causing training divergence. The fundamental issue is that independent factor updates provide no mechanism to control the spectral properties of the resultant product matrix $W$.

Existing approaches acknowledge this difficulty by maintaining dependencies on full-rank components. \citet{huh2024training} apply low-rank constraints only to adapter modules while keeping backbone weights full-rank. GaLore~\citep{zhao2024galore} and \citet{chenfira} project gradients to low-rank subspaces but maintain full-rank weights during training, thus reducing only optimizer state memory. \citet{wang2021pufferfish} employ hybrid architectures by converting later layers to low rank while keeping initial layers full-rank, and they initialize from checkpoints of models pretrained with full-rank weights to avoid early instabilities. Most closely related, \citet{wei2024building} restrict low-rank factorization to feed-forward layers and require full-rank initialization through self-guided training before transitioning to factorized weights. These methods avoid core instability by never fully committing to end-to-end factorized training.


We address this pathology through \textbf{Spect}ral renormalization combined with orthogonalizati\textbf{on} that directly targets the instability mechanism. Our method \textbf{Spectron} enables stable native low-rank training without auxiliary full-rank components by orthogonalizing gradient updates and constraining them to a region bounded by the inverse of the sum of spectral norms of the low-rank factors. We implement this constraint efficiently via power iteration-based~\cite{vogels2019powersgd} spectral estimation and Newton Schulz~\cite{jordan2024muon} based gradient orthogonalization that has negligible computational overhead, provably limiting spectral norm growth of the product matrix. This prevents the unbounded singular value expansion that destabilizes low-rank optimization.


We make the following contributions:
\begin{itemize}
\item \textbf{Spectral renormalization and orthogonalization:} We propose Spectron, an adaptive spectral norm constrained low-rank factor update with orthogonalization. We show that it bounds the resultant weight update spectral norms, enabling stable end-to-end factorized training from random initialization without auxiliary dense components (Section~\ref{sec:spectral_renormalization}).

\item \textbf{Empirical validation and scaling properties:} We demonstrate that our method applied to standard LLM pretraining with factorized non-embedding matrices enables stable training and achieves better final perplexity and downstream task accuracies compared to baselines. We show that factorized models trained with Spectron exhibit favorable scaling properties, matching or exceeding dense model performance (Figure~\ref{fig:flops_teaser}) across three model scales on FineWeb~\cite{penedo2024fineweb} pretraining (Section~\ref{sec:experiments}).



\item \textbf{Compute optimal factorized transformers:} We derive compute-optimal scaling laws for low-rank transformers through isoFLOP analysis across 47M--1.5B parameters and 250M--90B tokens, establishing scaling relationships analogous to Chinchilla laws~\citep{hoffmann2022an} with exponents $N_{\text{opt}} \propto C^{0.479}$ and $D_{\text{opt}} \propto C^{0.521}$ (Section~\ref{sec:scaling_laws}).
\end{itemize}

\begin{figure}[t!]
    \centering
    \includegraphics[width=0.9\linewidth]{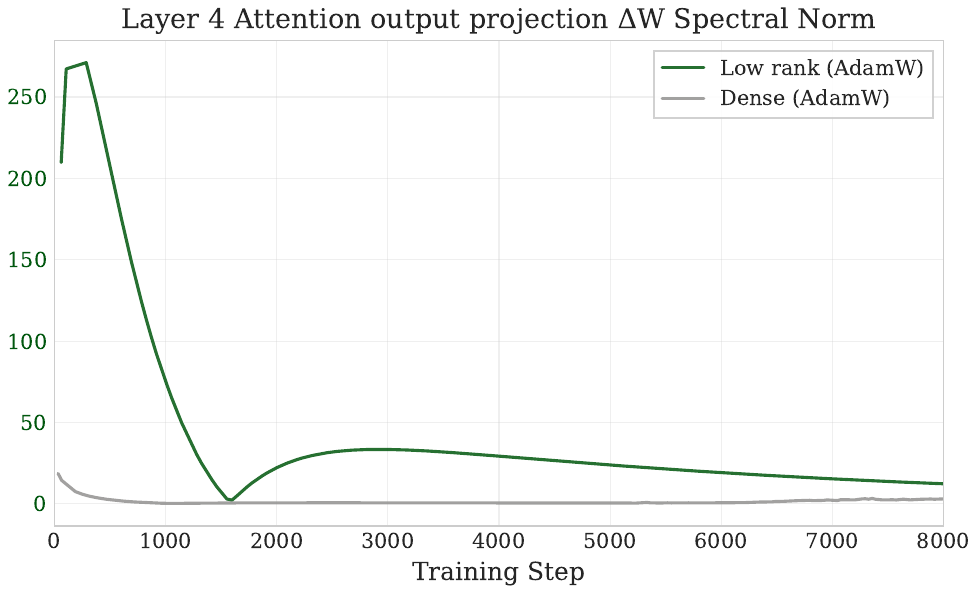}
    \caption{\textbf{Low-Rank Parameterization Destabilizes Spectral Norm Dynamics.} Weight update spectral norm ($\norm{\Delta W}_{2}$) comparison between low-rank (\textcolor{green}{green}) and dense (\textcolor{gray}{gray}) AdamW~\cite{kingma2015adam} training on layer 4 attention output projection of a Transformer~\cite{vaswani2017attention}. Dense training maintains stable, bounded spectral norms, while low-rank factorization exhibits 10-30$\times$ higher spectral norm magnitudes, revealing that the factorized updates (Equation~\eqref{eqn:chain_rule}) fundamentally cause spectral instability.}
    \label{fig:unstable_training}
    \vspace{-6mm}
\end{figure}

\section{Related Works}
\textbf{Low-Rank Factorization in Neural Networks}
Deep neural networks are shown to inherently develop low-rank representations during training. Early work on network pruning demonstrated that low-rank structures can approximate trained models with minimal performance loss~\cite{lecun1989optimal,yu2017compressing}, later attributed to implicit regularization in optimization dynamics~\cite{martin2021implicit,yang2023spectral,franklelottery}. \cite{pmlr-v280-galanti25a} formalized this by proving that SGD with weight decay induces rank collapse in weight matrices.
This observation motivated explicit low-rank parameterizations. Researchers have explored factorized architectures in both convolutional and transformer models~\cite{khodakinitialization,wei2024investigating}, while practitioners leverage low-rank structure to reduce fine-tuning costs through LoRA and its variants~\cite{hu2022lora,dettmers2023qlora,sharmatruth}.
Applying low-rank methods to pretraining presents greater challenges. \citet{huh2024training} investigates LoRA-style adapters by freezing full-rank weights during training, while gradient compression methods exploit low-rank structure for memory efficiency~\cite{zhao2024galore,chenfira,lialin2024relora}. Hybrid approaches like \cite{wang2021pufferfish} initialize from pretrained full-rank models to avoid optimization difficulties.
\citet{qiu2024compute} study factorized pretraining through the lens of $\mu$P~\citep{yang_tensor_2022}, deriving scaling laws for structured layers and showing that such layers can outperform dense layers at a fixed compute budget. In the Bayesian setting, \citet{toure2026singular} exploit low-rank factorization to concentrate the posterior on a rank-$r$ manifold, tightening PAC-Bayes bounds and match the baseline performance. 

Most closely related, \citet{wei2024building} solely factorizes fully connected layers while maintaining auxiliary full-rank weights for optimization stability. Our method differs by converting all non-embedding matrices to low-rank parameterizations without auxiliary weights, eliminating memory overhead while directly addressing optimization challenges in native low-rank training.\\
\textbf{Gradient Orthogonalization}
Gradient orthogonalization has emerged as a powerful technique for enhancing sample efficiency and optimization stability in deep neural networks \citep{jordan2024muon,bernstein2024old,bernstein2025deriving,ma2024swan,pethick2025training}. The Muon optimizer and its variants \citet{jordan2024muon,ahn2025dion,si2025adamuon,pethick2025training} preprocess gradients through orthogonalization, achieving faster convergence and improved training dynamics\cite{vasudeva2025muon}. Recent work demonstrates that this approach scales effectively to large language model training \cite{liu2025muon,kimiteam2025kimik2openagentic}.
Several works~\cite{kovalev2025understanding,bernstein2025deriving,chen2025muon,li2025note,fan2025implicit} show that orthogonalized updates perform weight updates under spectral norm constraints. \citet{wei2024building} observes uncontrolled spectral norm growth in weight matrices of the factors as a primary source of instability causing large gradient norms when training low-rank factorized networks. Yet they failed to see the effect on the resultant product matrix and opted to use dense guidance as the solution. We directly address this challenge by employing gradient orthogonalization and spectral renormalization to regulate spectral norm growth throughout training.

\section{Background and Problem Formulation}
\label{sec:background_problem}

We establish the foundational concepts underlying our approach and formalize the spectral instability challenge inherent to low-rank factorized training.
\subsection{Background}
\textbf{Low-Rank Parameterization}

We parameterize non-embedding layer weight matrices $W \in \mathbb{R}^{m \times n}$ of a transformer~\cite{vaswani2017attention} neural network $f_\theta$ using low-rank factorizations to minimize computational overhead during training with next-token prediction cross-entropy loss $\mathcal{L}$. Specifically, we represent $W$ as:
\begin{equation}
    W = AB^\top, \quad A \in \mathbb{R}^{m \times r},\ B \in \mathbb{R}^{n \times r},
    \label{eq:low_rank_def}
\end{equation}
where $r$ denotes the rank with $r < \min(m,n)$. 

During training, gradient-based updates are applied directly to the factors $A$ and $B$. Via the chain rule, this leads to the composite weight update:
\begin{equation}
\label{eqn:chain_rule}
\Delta W = \Delta A\, B^\top + A\, \Delta B^\top + \Delta A\, \Delta B^\top
\end{equation}

\textbf{Operator Norms and Spectral Stability}

We utilize the spectral norm $\norm{W}_{2}$ to measure training stability, defined as the largest singular value of $W$:
\begin{equation}
    \norm{W}_{2} = \sup_{x \in \mathbb{R}^n \setminus \{0\}} \frac{\abs{Wx}_2}{\abs{x}_2},
\end{equation}
where $\abs{}_2$ denotes the Euclidean norm of the vector. The spectral norm satisfies the submultiplicative property
\begin{equation}
    \norm{XY}_{2} \leq \norm{X}_{2} \norm{Y}_{2},
    \label{eqn:submultiplicativity}
\end{equation}
which proves essential for bounding composite updates in factorized layers. Following \citet{bernstein2025deriving} and \citet{yang2023spectral}, we employ the Root Mean Square (RMS) norm for a vector $y \in \mathbb{R}^m$:
\begin{equation}
    \abs{y}_{\mathrm{rms}} = \sqrt{\frac{1}{m} \sum_{i=1}^m y_i^2},
\end{equation}
and the RMS-to-RMS operator norm for a matrix $W$, which measures the maximum amplification of entry-wise magnitudes:
\begin{equation}
    \norm{W}_{\mathrm{rms} \to \mathrm{rms}} = \sup_{x \in \mathbb{R}^n \setminus \{0\}} \frac{\abs{Wx}_{\mathrm{rms}}}{\abs{x}_{\mathrm{rms}}}.
\end{equation}
This relationship enables us to control activation variance by constraining the spectral norm of weight updates.

\textbf{Gradient Orthogonalization}

Gradient orthogonalization has emerged as a principled technique for accelerating neural network training by constraining the geometry of hidden layer matrix parameter updates~\cite{kovalev2025understanding,bernstein2025deriving,liu2025muon,jordan2024muon}. The core principle normalizes all singular values of the update to unity.

Formally, given a gradient matrix $G_t = \nabla_\theta \mathcal{L}$ at time step $t$ with singular value decomposition $G_t = U\Sigma V^\top$, we define the orthogonalization operation $\mathrm{Ortho}$ as:
\begin{equation}
    O_t = \mathrm{Ortho}(G_t) = UV^\top.
\end{equation}
In practice, \citet{jordan2024muon} orthogonalize the updates from SGD with momentum ($M_t$) using efficient Newton--Schulz iterations (Algorithm~\ref{alg:newton_schulz}), yielding the update rule:
\begin{equation}
    \theta_t \leftarrow \theta_{t-1} - \eta \cdot O_t,
\end{equation}
where $\eta$ denotes the learning rate.

\subsection{The Spectral Instability Problem in Low-Rank Training}

The optimization difficulties in low-rank training stem from uncontrolled growth in the spectral norm of the weight matrix updates. We note that this aligns with the observations of ~\citet{wei2024building} where the authors found a correlation between large gradient norms and high spectral norms of the factors. Following the analysis of \citet{bernstein2025deriving}, consider the activation $y = Wx$ and its induced change under a weight update:
\begin{equation}
    \Delta y = \Delta W\, x.
\end{equation}

The change in activation can be bounded using the RMS-to-RMS operator norm:
\begin{equation}
    |\Delta y|_{\mathrm{rms}} \leq \norm{\Delta W}_{\mathrm{rms} \to \mathrm{rms}} = \sqrt{\frac{n}{m}} \|\Delta W\|_{2}
\end{equation}

When the spectral norm $\|\Delta W\|_{2}$ grows excessively, the resulting activation changes destabilize training. Interestingly, this phenomenon is exclusive to low-rank factorized training and does not manifest in dense model training. Figure~\ref{fig:unstable_training} demonstrates this contrast: dense training with AdamW~\cite{kingma2015adam} maintains stable, bounded spectral norms throughout optimization, while low-rank factorization exhibits 10-30$\times$ higher spectral norm magnitudes. This reveals that the factorized update structure in Equation~\eqref{eqn:chain_rule} fundamentally causes spectral instability.

\begin{figure*}[!]
    \centering
    
    \begin{subfigure}{0.33\textwidth}
        \centering
        \includegraphics[width=\textwidth]{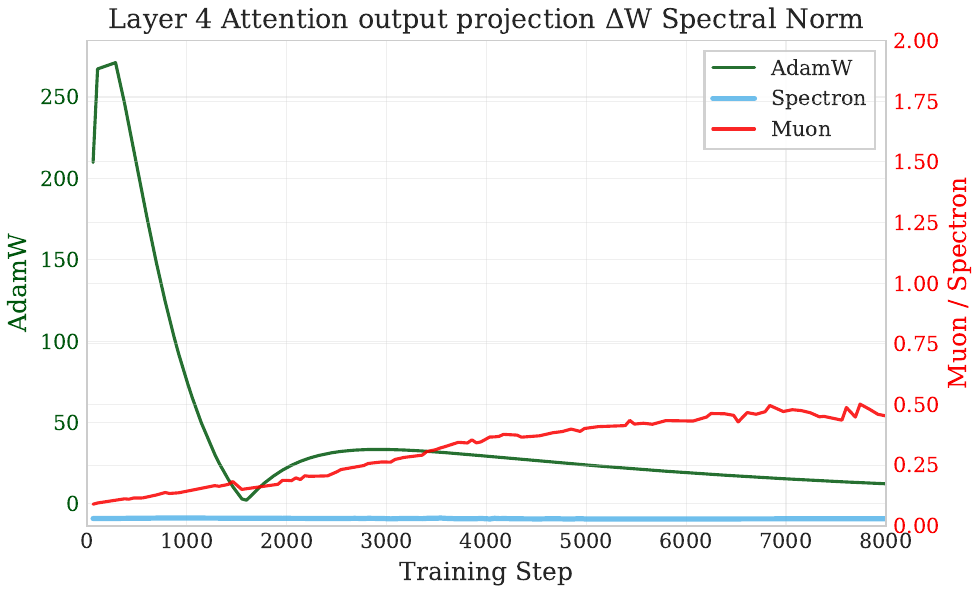}
        \caption{Variation of $||\Delta W||_{2}$ with training step}
        \label{fig:delta_w_spectral_norm}
    \end{subfigure}%
    \hfill
    \begin{subfigure}{0.33\textwidth}
        \centering
        \includegraphics[width=\textwidth]{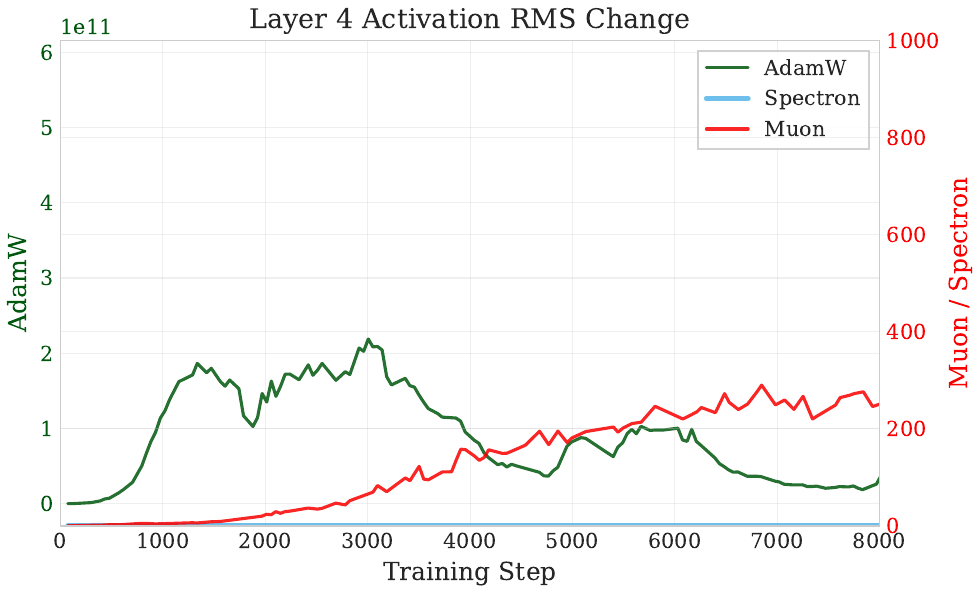}
        \caption{Variation of $|\Delta y|_{rms}$ with training step}
        \label{fig:rms_activation}
    \end{subfigure}%
    \hfill
    \begin{subfigure}{0.33\textwidth}
        \centering
        \includegraphics[width=\textwidth]{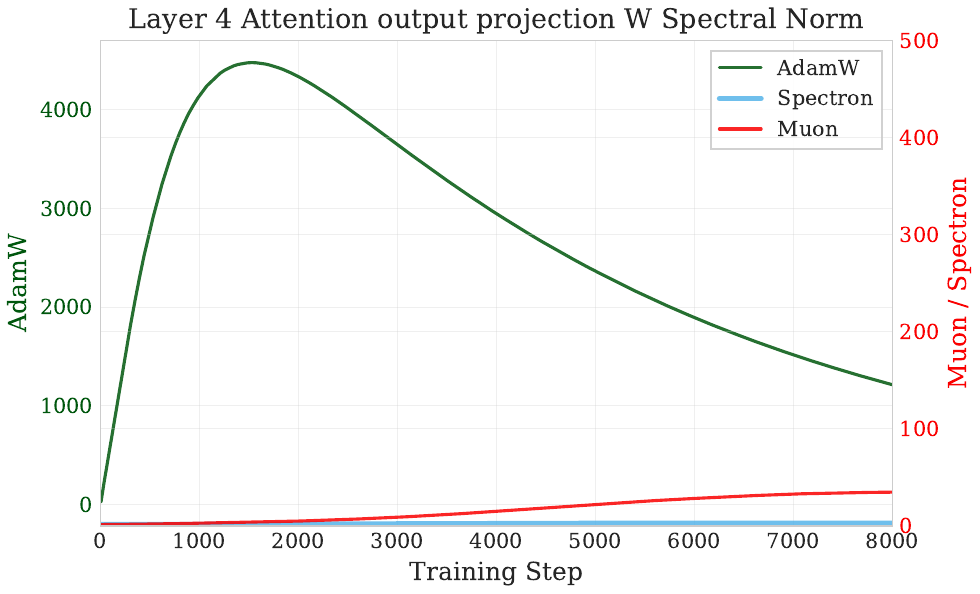}
        \caption{Variation of $||W||_{2}$ with training step}
        \label{fig:spectral_norm_change}
    \end{subfigure}
    
    \caption{\textbf{Spectral Norm Constraints Stabilize Low-Rank Training.} Comparison of (a) weight update spectral norm $\norm{\Delta W}_{2}$, (b) activation RMS change $\abs{\Delta y}_{rms}$, and (c) weight spectral norm $\norm{W}_{2}$ across 8000 training steps for layer 4 attention output projection of a 94M parameter Factorized Transformer~\citep{vaswani2017attention}. AdamW~\cite{kingma2015adam} (\textcolor{green}{green}, left axis) exhibits explosive growth in all metrics with unconstrained spectral norm dynamics. Muon~\citep{jordan2024muon} (\textcolor{red}{red}, right axis) achieves moderate control through gradient orthogonalization~\citep{bernstein2024old}. Our method, Spectron (\textcolor{blue}{blue}, right axis) maintains bounded spectral norms throughout training by adaptively constraining factor updates, demonstrating stable optimization. Note that AdamW curves use a different y-axis scale (left) compared to Muon and Spectron (right) for visualization purposes.}
    \label{fig:spectral_norm_comparison}
\end{figure*}
\vspace{-3mm}


\begin{algorithm}[t!]
\small
\caption{\textbf{Spectron}}
\label{alg:spectral_scaling_momentum}
\begin{algorithmic}[1]
\Require Weight matrices $A \in \mathbb{R}^{m \times r}$, $B \in \mathbb{R}^{n \times r}$, step size $\eta > 0$
\Require Momentum decay $\beta \in [0,1)$ (e.g. 0.9 or 0.95)
\Require Number of power iteration steps $k_{\text{power}}$ (default: 1), Newton-Schulz iterations $k_{\text{ns}}$ (default: 5)
\Statex
\State Initialize $u_A \in \mathbb{R}^{m}$ and $u_B \in \mathbb{R}^{n}$ randomly and normalize: $u_A \leftarrow u_A / \|u_A\|$, $u_B \leftarrow u_B / \|u_B\|$
\State Initialize momentum buffers $M_A \leftarrow 0 \in \mathbb{R}^{m \times r}$, $M_B \leftarrow 0 \in \mathbb{R}^{n \times r}$
\State $t \leftarrow 1$
\While{not converged}
    \State $G_A^{(t)} \leftarrow \nabla_A \mathcal{L}$ \Comment{Gradient of loss w.r.t. $A$}
    \State $G_B^{(t)} \leftarrow \nabla_B \mathcal{L}$ \Comment{Gradient of loss w.r.t. $B$}
    
    \State $M_A \leftarrow \beta \, M_A + (1-\beta) \, G_A^{(t)}$ \Comment{Momentum update}
    \State $M_B \leftarrow \beta \, M_B + (1-\beta) \, G_B^{(t)}$ \Comment{Momentum update}
    
    \State $O_A^{(t)} \leftarrow \text{Ortho}(M_A, k_{\text{ns}})$ \Comment{(Algorithm \ref{alg:newton_schulz})}
    \State $O_B^{(t)} \leftarrow \text{Ortho}(M_B, k_{\text{ns}})$ \Comment{(Algorithm \ref{alg:newton_schulz})}
    
    \State $\sigma_A, u_A \leftarrow \text{PowerIter}(A^{(t)}, u_A, k_{\text{power}})$ \Comment{(Algorithm \ref{alg:power_iter})}
    \State $\sigma_B, u_B \leftarrow \text{PowerIter}(B^{(t)}, u_B, k_{\text{power}})$ \Comment{(Algorithm \ref{alg:power_iter})}
    
    \State $\Delta_A \leftarrow \frac{\eta}{\sigma_A + \sigma_B + 1} \cdot O_A^{(t)}$
    \State $\Delta_B \leftarrow \frac{\eta}{\sigma_A + \sigma_B + 1} \cdot O_B^{(t)}$
    
    \State $A^{(t)} \leftarrow A^{(t-1)} - \Delta_A$
    \State $B^{(t)} \leftarrow B^{(t-1)} - \Delta_B$
    
    \State $t \leftarrow t + 1$
\EndWhile
\State \Return $A^{(t)}, B^{(t)}$
\end{algorithmic}
\end{algorithm}

\section{Spectron: Spectral Renormalization and Orthogonalization}
\label{sec:spectral_renormalization}

Drawing on recent works establishing orthogonalized updates as updates under a spectral norm constraint~\cite{kovalev2025understanding,bernstein2025deriving,bernstein2020distance,chen2025muon,li2025note,fan2025implicit}, we propose to control training stability by bounding the spectral norm of the composite update:
\begin{equation}
    \|\Delta W\|_{2} \leq \eta,
\end{equation}
where $\eta$ is a prescribed constraint radius defined by the learning rate. Our approach constrains the magnitudes of factor updates $\Delta A$ and $\Delta B$ such that the composite update $\Delta W$ maintains a stable spectral norm.

\textbf{Bounding $\Delta W$ Through Adaptive Factor Constraints}

We leverage gradient orthogonalization to ensure that factor updates remain within the local constraint radius defined by spectral norm~\cite{kovalev2025understanding}:
\begin{equation}
    \|\Delta A\|_{2} \leq \rho, \qquad \|\Delta B\|_{2} \leq \rho
    \label{eqn:bounds}
\end{equation}
for a constraint radius $\rho$ to be determined adaptively. Applying the triangle inequality and submultiplicativity of the spectral norm to Equation~\eqref{eqn:chain_rule}:
\begin{equation}
    \|\Delta W\|_{2} \leq \|\Delta A\, B^\top\|_{2} + \|A\, \Delta B^\top\|_{2} + \|\Delta A\, \Delta B^\top\|_{2}.
\end{equation}

Using property~\eqref{eqn:submultiplicativity} and bounds in Equation~\eqref{eqn:bounds}:
\begin{equation}
    \|\Delta W\|_{2} \leq \rho\,\|B\|_{2} + \rho\,\|A\|_{2} + \rho^2.
\end{equation}

For typical learning rates where $\rho < 1$, we obtain the upper bound:
\begin{equation}
    \|\Delta W\|_{2} \leq \rho\bigl(\|A\|_{2} + \|B\|_{2} + 1\bigr).
\end{equation}

To satisfy the global constraint $\|\Delta W\|_{2} \leq \eta$, we dynamically set the local constraint radius as:
\begin{equation}
    \rho = \frac{\eta}{\|A\|_{2} + \|B\|_{2} + 1}.
\label{eqn:final_bound}
\end{equation}

This adaptive scaling ensures that orthogonalized updates $(\Delta A, \Delta B)$ satisfying Equation~\eqref{eqn:final_bound} induce a composite update $\Delta W$ that respects the required spectral norm bound, regardless of the current magnitudes of $A$ and $B$. We estimate $\|A\|_{2}$ and $\|B\|_{2}$ efficiently using a single power iteration (Algorithm~\ref{alg:power_iter}) and orthogonalize using five Newton Schulz iterations (Algorithm~\ref{alg:newton_schulz}) , following recent works on efficient spectral norm approximation~\citep{ahn2025dion,vogels2019powersgd} and orthogonalization~\citep{jordan2024muon}. In practice, Spectron can be implemented by combining  orthogonalized updates from Muon~\citep{jordan2024muon} for each factor with an explicit spectral renormalization step based on the estimated norms of both factors. The complete algorithm of \textbf{Spectron} is presented in Algorithm~\ref{alg:spectral_scaling_momentum}. 



\textbf{Empirical Validation of Spectral Constraints}

Figure~\ref{fig:spectral_norm_comparison} demonstrates the effectiveness of our spectral renormalization approach across 8000 training steps for layer 4 attention output projection of a 94M parameter factorized Transformer~\citep{vaswani2017attention}. Standard AdamW~\cite{kingma2015adam} exhibits explosive growth in weight update spectral norm $\norm{\Delta W}_{2}$, activation RMS change $\abs{\Delta y}_{rms}$, and weight spectral norm $\norm{W}_{2}$, confirming the instability identified in Section~\ref{sec:background_problem}. The orthogonalized optimizer Muon~\citep{jordan2024muon} achieves moderate control through gradient orthogonalization alone~\citep{bernstein2024old}. Our method maintains bounded spectral norms throughout training by adaptively constraining factor updates according to Equation~\eqref{eqn:final_bound}, demonstrating stable optimization dynamics across all tracked metrics.


\begin{figure}
    \centering
    \includegraphics[width=0.9\linewidth]{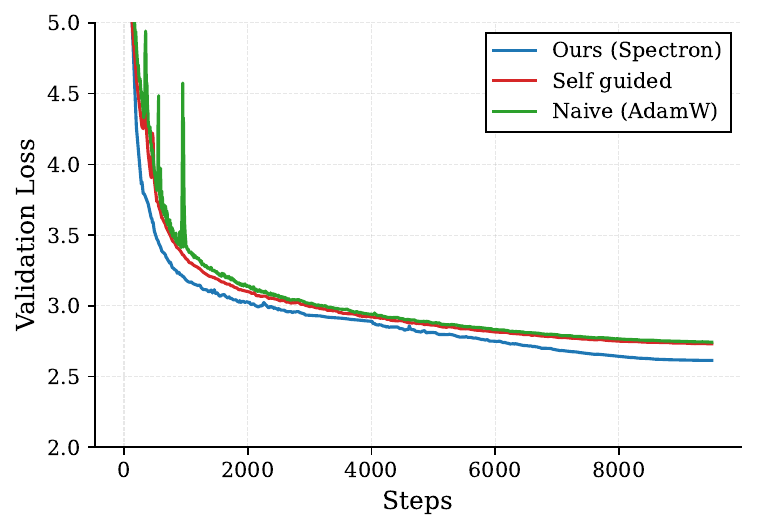}
    \caption{\textbf{Spectrally Normalized Low-Rank Training Outperforms Baselines.} Validation loss on FineWeb~\cite{penedo2024fineweb} held-out set during Factorized Transformer-M (297M) pretraining comparing Spectron (\textcolor{blue}{blue}), self-guided training (\textcolor{red}{red}), and naive AdamW (\textcolor{green}{green}). Our approach achieves both faster initial convergence and superior final performance (Table~\ref{tab:low-rank-comparison}), outperforming self-guided training despite its dense auxiliary full rank weights, while maintaining sub-1\% computational overhead compared to self-guided's 25\% additional FLOPs.}
    \label{fig:self-guided-compare}
\end{figure}

\begin{table*}[h!]
\small
\setlength{\tabcolsep}{2pt}
\centering
\begin{tabular}{@{} l r r r r @{}}
\toprule
\textbf{Method}                                           & \multicolumn{1}{l}{\textbf{Perplexity ($\downarrow$)}} & \multicolumn{1}{l}{\textbf{HellaSwag ($\uparrow$)}} & \multicolumn{1}{l}{\textbf{PIQA ($\uparrow$)}} & \multicolumn{1}{l}{\textbf{Arc Easy ($\uparrow$)}} \\ \midrule
\multicolumn{5}{c}{Factorized Llama-94M}                                                                                                                                                                       \\ \midrule
Naive (AdamW)~\cite{kingma2015adam} & 26.43                                   & 26.83                                  & 57.83                             & 31.44                                 \\
Self guided~\cite{wei2024building}  & 24.17                                   & 26.62                                  & 58.32                             & 30.93                                 \\
Ours (Spectron)                                           & \textbf{21.86}                          & \textbf{27.52}                         & \textbf{58.60}                    & \textbf{31.90}                        \\ \midrule
\multicolumn{5}{c}{Factorized Llama-297M}                                                                                                                                                                      \\ \midrule
Naive (AdamW)~\cite{kingma2015adam} & 15.54                                   & 31.59                                  & 63.00                             & 34.09                                 \\
Self guided~\cite{wei2024building}  & 15.53                                   & 31.08                                  & 62.89                             & 34.05                                 \\
Ours (Spectron)                                           & \textbf{14.62}                          & \textbf{33.98}                         & \textbf{63.55}                    & \textbf{35.10}                        \\ \midrule
\multicolumn{5}{c}{Factorized Llama-454M}                                                                                                                                                                      \\ \midrule
Naive (AdamW)~\cite{kingma2015adam} & 14.57                                   & 34.05                                  & 65.18                             & 35.98                                 \\
Self guided~\cite{wei2024building}  & 13.70                                   & 34.85                                  & 64.91                             & 35.82                                 \\
Ours (Spectron)                                           & \textbf{12.11}                          & \textbf{40.11}                         & \textbf{66.76}                    & \textbf{36.78}                        \\ \bottomrule
\end{tabular}
\vspace{5mm}
\caption{\textbf{Comparative Performance of Low-Rank Factorized Training Methods Across Model Scales.} Perplexity ($\downarrow$), and downstream task accuracies (HellaSwag~\cite{zellers2019hellaswag} ($\uparrow$), PIQA~\cite{bisk2020piqa} ($\uparrow$), Arc Easy~\cite{clark2018think} ($\uparrow$)) for three factorized Transformer variants (94M, 297M, 454M parameters) trained with naive AdamW~\cite{kingma2015adam}, self-guided training~\cite{wei2024building}, and our method Spectron. Our approach consistently achieves superior performance across all metrics and model sizes.}
\label{tab:low-rank-comparison}
\end{table*}
\section{Experiments}
\label{sec:experiments}

\begin{figure}
    \centering
    \includegraphics[width=\linewidth]{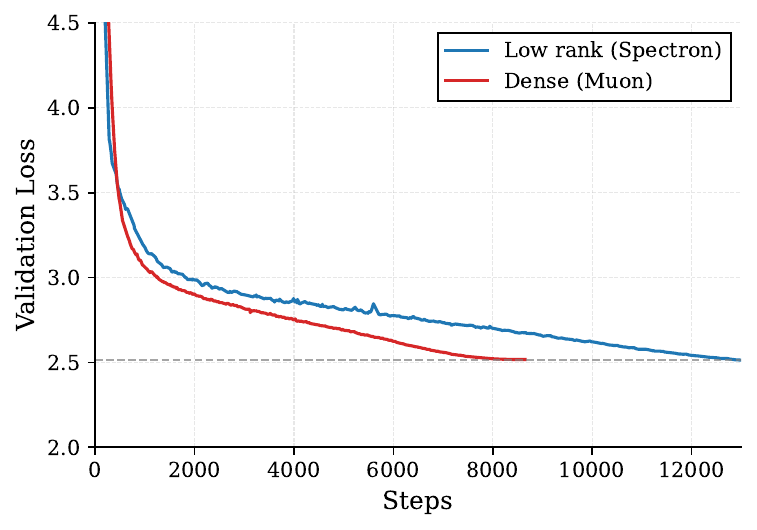}
    \caption{\textbf{Low-Rank Factorization Matches Dense Performance with Longer Training.} Validation loss comparison between Dense Transformer-L (780M parameters) and our Low-Rank Factorized Transformer-L (454M parameters) trained for \textbf{equal FLOPs} by matching training steps. Despite a $\sim42\% $ parameter reduction, our factorized model (\textcolor{blue}{blue}) converges to the same final validation loss as the dense baseline (\textcolor{red}{red}), demonstrating that compute-equivalent training yields an inference-optimal model.}
    \label{fig:dense_compare_training}
\end{figure}

\textbf{Experimental Setup}
We evaluate Spectron on LLaMA-style transformer architectures~\cite{touvron2023llama} across multiple scales. We train three full-rank variants: Llama-134M, Llama-500M and Llama-780M, alongside corresponding factorized versions with rank ratio 0.25 ($r=0.25 n$): Factorized Llama-94M, Factorized Llama-297M, and Factorized Llama-454M. All non-embedding weight matrices use low-rank decompositions in factorized models (implementation details in Appendix~\ref{apdx:implementation_details}).

All models are pretrained on FineWeb~\cite{penedo2024fineweb} with a 100M token validation set. Dense baselines train to Chinchilla~\cite{hoffmann2022an}-optimal token counts and corresponding factorized models train for matched FLOPs. We train the dense baselines with Muon~\cite{jordan2024muon} optimizer for fair comparison. We report validation perplexity, and normalized accuracy on HellaSwag~\cite{zellers2019hellaswag}, PIQA~\cite{bisk2020piqa}, and ARC-easy~\cite{clark2018think} via \texttt{lm-evaluation-harness}~\cite{eval-harness}.\\
\paragraph{Baselines.}
We compare against self-guided training~\cite{wei2024building}, the current state-of-the-art for stable low-rank pretraining. This method supervises low-rank parameters with concurrent dense weight updates during the first half of training:
While effective, this incurs $\sim$25\% FLOP overhead during guidance (Appendix~\ref{apn:self-guided}). We additionally benchmark naive AdamW\cite{kingma2015adam} training.

Our method introduces minimal overhead: Newton-Schulz  orthogonalization~\cite{jordan2024muon} (Algorithm~\ref{alg:newton_schulz}) adds $6k_{\text{ns}}nm^2$ FLOPs ($<$1\% for typical architectures), while power iteration spectral norm estimation (Algorithm~\ref{alg:power_iter}) requires only $2mn$ FLOPs per matrix of size $m\times n$. Total overhead remains sub-1\%---a $25\times$ reduction versus self-guided training.

\subsection{Comparison to Low rank training baselines}

Figure~\ref{fig:self-guided-compare} demonstrates that spectron enables both faster convergence and superior final performance compared to existing low-rank training methods. Our approach achieves stable training at higher learning rates that cause baseline divergence (Appendix~\ref{apdx:ablation_stability}), indicating effective constraint of spectral norm dynamics. Table~\ref{tab:low-rank-comparison} confirms consistent improvements across model scales: Spectron reduces perplexity  by 6--12\% versus self-guided training and 6--17\% versus naive AdamW, with corresponding downstream accuracy gains.

\subsection{Comparison to Dense Model Training}

To evaluate whether low-rank factorization inherently limits model capacity, we compare factorized transformers against dense baselines trained with equal computational budgets.

\begin{figure}[t!]
    \centering
    

    \includegraphics[width=0.99\linewidth]{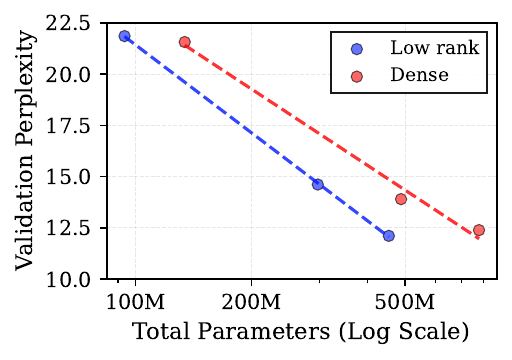}
    \caption{\textbf{Low-Rank Factorization Improves Scaling Efficiency.} Validation perplexity($\downarrow$) comparison between low-rank factorized models (\textcolor{blue}{blue}) and dense models (\textcolor{red}{red}) across model sizes from 100M to 780M parameters. Low-rank models achieve consistently lower perplexity across scales and need a lower parameter count for a given perplexity, showing inference efficiency}
    \label{fig:dense_compare_val_perplexity}
    
    \vspace{1em} 
    
    \includegraphics[width=0.99\linewidth]{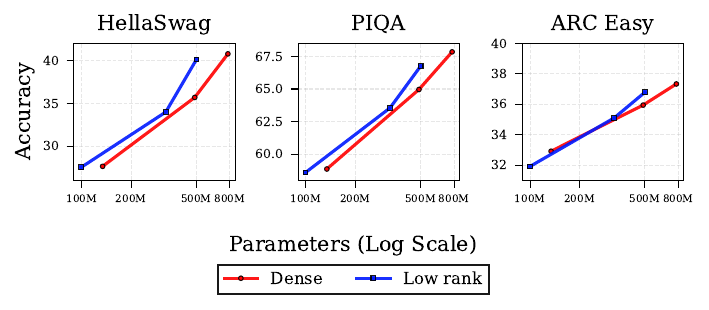}
    \caption{\textbf{Low-Rank Models Achieve Superior Downstream Performance with Fewer Parameters:} Accuracy($\uparrow$) comparison across three benchmark tasks (HellaSwag~\cite{zellers2019hellaswag}, PIQA~\cite{bisk2020piqa}, ARC Easy~\cite{clark2018think}) for dense versus low-rank models trained with equal computational budgets. Low-rank architectures (\textcolor{blue}{blue}) consistently outperform dense baselines (\textcolor{red}{red}) across all model scales, demonstrating strong downstream performance under reduced inference cost}
    \label{fig:downstream_scatter}
    
\end{figure}

Figure~\ref{fig:flops_teaser} and ~\ref{fig:dense_compare_training} shows that when trained for equal FLOPs, Factorized Transformer-L (454M) converges to the same validation loss as Dense Transformer-L (780M), demonstrating that low-rank capacity limitations can be overcome through extended training under stable optimization~\citep{sardana2024beyond}. Even with 42\% reduced parameters, we recover the same performance, yielding inference savings by the same amount. This phenomenon strengthens at scale: Figure~\ref{fig:dense_compare_val_perplexity} reveals that factorized models achieve consistently lower perplexity than parameter-matched dense baselines. This yields a more compact model for a given perplexity threshold, thereby substantially reducing inference costs. 

Figure~\ref{fig:downstream_scatter} extends these findings to downstream evaluation, where factorized transformers match or exceed dense performance across HellaSwag~\cite{zellers2019hellaswag}, PIQA~\cite{bisk2020piqa}, and ARC Easy~\cite{clark2018think} benchmarks. 

\begin{figure*}[t!]
    \centering
    \includegraphics[width=0.99\textwidth]{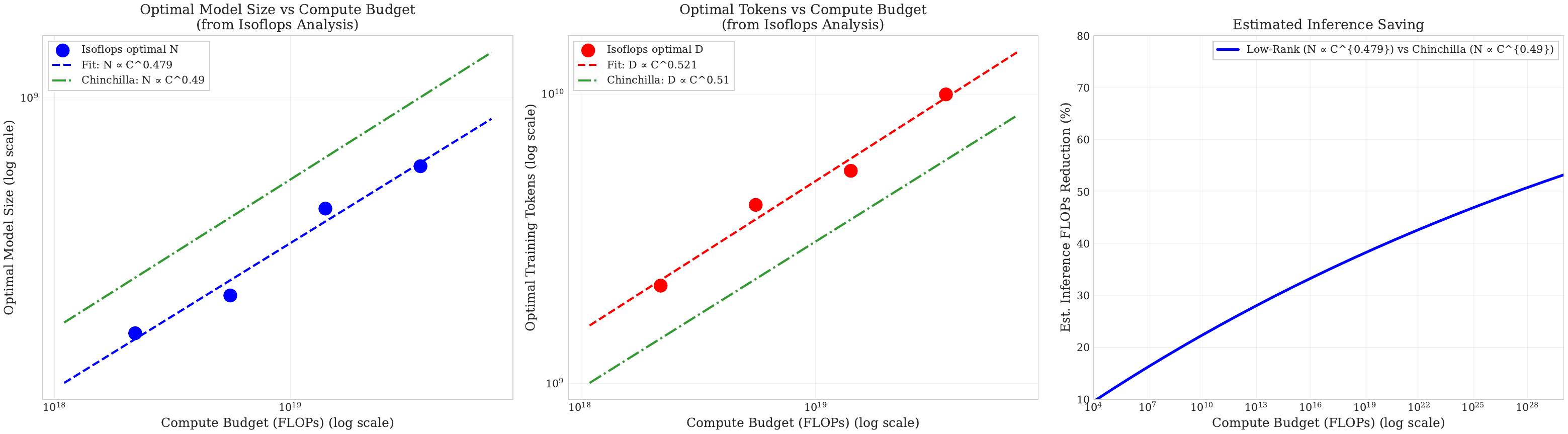}
    \caption{\textbf{Factorized Models Scale More Conservatively Than Dense Transformers, Yielding Substantial Inference Efficiency Gains.} 
    (\textit{Left}) Optimal model size versus compute budget for low-rank architectures follows $N_{\text{opt}} \propto C^{0.479}$ (\textcolor{blue}{blue} points), compared to Chinchilla's $N \propto C^{0.49}$ (\textcolor{green}{green}  dashed reference). 
    (\textit{Center}) Optimal training tokens scale as $D_{\text{opt}} \propto C^{0.521}$ (\textcolor{red}{red} points), versus Chinchilla's $D \propto C^{0.51}$. 
    The reduced parameter scaling exponent ($0.479$ vs.\ $0.49$) indicates that compute-optimal low-rank models are smaller than their dense counterparts at equivalent training budgets, requiring proportionally more training tokens.
    (\textit{Right}) Estimated inference cost savings, computed as $(1 - N_{\text{opt}} / N_{\text{Chinchilla}}) \times 100 = (1 - 1/C^{0.011}) \times 100\%$, assuming inference cost scales as $2N_{\text{opt}} \cdot D_{\text{inf}}$ with identical proportionality constants for both low-rank and dense models. 
    Under contemporary FLOP budgets for training ($\sim 10^{26}$ FLOPs), low-rank models achieve up to 50\% inference cost reduction compared to Chinchilla-optimal dense transformers.
    This trade-off favors scenarios where inference efficiency and compact deployment are prioritized over token budget.}
    \label{fig:optimal_allocation}

\end{figure*}
\begin{figure}
    \centering
    \includegraphics[width=0.9\linewidth]{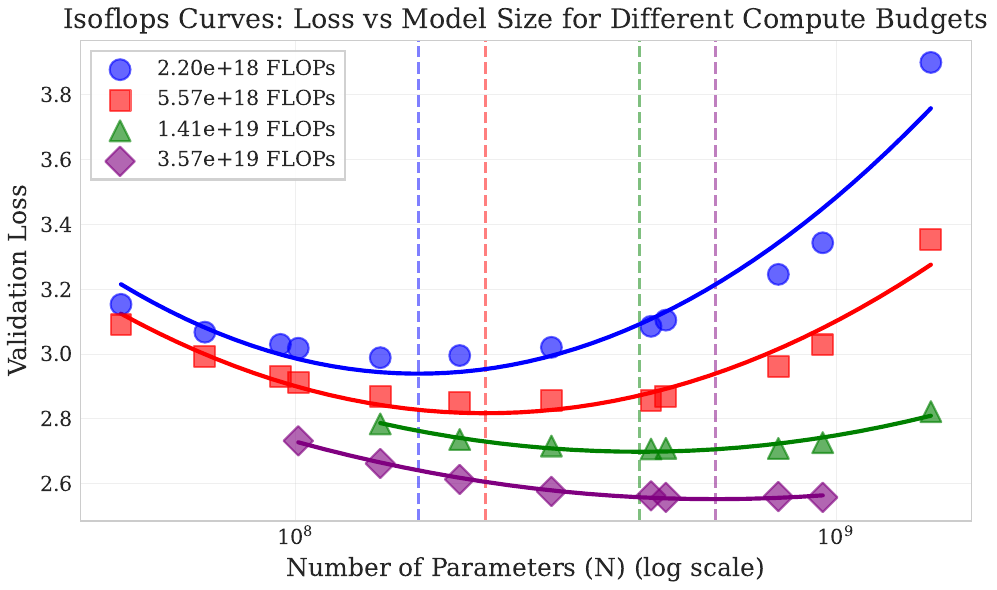}
    \caption{\textbf{Low-Rank Architectures Exhibit Clear Compute-Optimal Model Sizes.} Validation loss versus parameter count for factorized transformers trained at four compute budgets (2.20e+18 to 3.57e+19 FLOPs). Each IsoFLOP curve displays a distinct minimum (vertical dashed lines), with optimal model size increasing monotonically with compute budget. This replicates the fundamental structure of Chinchilla scaling laws~\cite{hoffmann2022an} in the low-rank regime, establishing that compute-optimal training requires balanced scaling of both model size and training duration.}
    \label{fig:isoflops}
    \vspace{-3em}
\end{figure}
\section{Towards compute optimal Low-Rank Pretraining}
\label{sec:scaling_laws}

Having established performance parity with dense models under equal compute (Section~\ref{sec:experiments}), we investigate the fundamental scaling properties of low-rank architectures: \textit{given a fixed computational budget, what is the optimal allocation between model parameters and training tokens?}

We adopt the IsoFLOP profiling approach from \citet{hoffmann2022an}, training factorized transformers ranging from 47M to 1.5B parameters across four compute budgets ($2.20 \times 10^{18}$ to $3.57 \times 10^{19}$ FLOPs). Token budgets are adjusted inversely to maintain constant FLOPs per configuration, with quadratic fits to each IsoFLOP curve identifying the loss-minimizing model size at each compute level.

Figure~\ref{fig:isoflops} demonstrates through 39 extensive pretraining runs that low-rank pretraining exhibits well-defined compute-optimal model sizes, with clear loss minima at each budget level. These optima shift rightward with increased compute, mirroring the scaling structure of dense transformers~\cite{hoffmann2022an} and confirming that low-rank architectures follow predictable optimization frontiers.
Figure~\ref{fig:optimal_allocation} reveals the precise relationships: optimal model size scales as $N_{\text{opt}} \propto C^{0.479}$ (versus Chinchilla's 0.49), while training tokens scale as $D_{\text{opt}} \propto C^{0.521}$ (versus 0.51). This modest deviation indicates that factorized architectures achieve compute-optimality at smaller model sizes, yielding inference-efficient models that compensate through extended training~\cite{sardana2024beyond}.

Low-rank pretraining naturally produces models that are simultaneously compute-optimal and inference-efficient. For a given compute budget, the resulting architecture contains fewer parameters than a comparably-trained dense model while maintaining equivalent performance(Section~\ref{sec:experiments}). This property is particularly advantageous when inference efficiency is prioritized in settings similar to what ~\citet{sardana2024beyond} explores, where data availability exceeds compute constraints, enabling full utilization of a large corpora.  A parametric study of this using Approach 3 of \citet{hoffmann2022an} appears in Appendix~\ref{apdx:approach_3}.

The near-equivalence of scaling exponents (0.479 vs. 0.49 for parameters; 0.521 vs. 0.51 for tokens) suggests that low-rank factorization does not fundamentally alter transformer scaling dynamics but rather shifts the compute-optimal frontier toward  smaller, more token-intensive configurations. But this is non-trivial at scale, as described in Figure~\ref{fig:optimal_allocation} (right), where at modern compute budgets this can be a significant reduction in the inference cost estimated by $2 N_{opt}D_{inf}$, where $D_{inf}$ denotes total inference tokens assumed to be the same for both configurations.

\section{Conclusion}
In this paper, we present Spectron, a method for training large language models (LLMs) from scratch using exclusively low-rank factorized weight matrices, eliminating the need for auxiliary full-rank weights. Our core contribution identifies unbounded spectral norm growth as the fundamental source of training instability in factorized architectures. We address this through a combination of spectral renormalization and gradient orthogonalization that adaptively constrains weight updates based on the spectral norm of factors. This approach bounds composite weight update norms while introducing negligible computational overhead relative to existing methods~\citep{wei2024building}.

Our empirical validation demonstrates that factorized transformers achieve performance parity with dense baselines under equivalent FLOP budgets, despite significantly fewer parameters. Through systematic IsoFLOP analysis, we derive compute-optimal scaling laws specific to native low-rank pretraining. The resulting relationships yield optimal model size scaling as $N_{\text{opt}} \propto C^{0.479}$ and training token requirements as $D_{\text{opt}} \propto C^{0.521}$, revealing that compute-optimal factorized architectures favor smaller model configurations trained on proportionally larger datasets. This scaling behavior translates directly to substantial inference-time compute savings. Future works could develop communication strategies specifically tailored for factorized architectures to reduce distributed training overhead. We primarily focus on large language models; the underlying principles suggest natural extensions to multimodal architectures. Finally, we anticipate that native low-rank training could enable more flexible pretraining paradigms.
\paragraph{Limitations.}
Our experiments are conducted on models up to 1.5B parameters which remain substantially smaller than contemporary foundation model pretraining regimes that often operate at trillion-token scale. Consequently, while the empirical results demonstrate consistent trends within the studied regime, direct validation at frontier scales remains an open question. Nevertheless, the fitted scaling laws suggest that the observed behavior may continue to hold at larger scales, providing preliminary evidence for the broader applicability of our conclusions. In addition, we observe wall-clock time gap between dense and low-rank training in our current implementation. Dense training benefits from highly optimized fused kernels, whereas our current low-rank implementation relies on two unfused matrix multiplications whose kernel launch overhead dominates at the scales considered here. We believe that custom fused kernels can substantially reduce these inefficiencies.

\section*{Acknowledgement}
EB and PJ acknowledge funding from FRQNT (\href{https://doi.org/10.69777/2002414}{https://doi.org/10.69777/2002414}) and NSERC. EO  acknowledges funding from PEPR IA (grant SHARP ANR-23-PEIA-0008). We acknowledge compute resources from IDRIS under the allocation 2025-AD011015884R1, CFI grant 43694 and Digital Research Alliance of Canada.


\section*{Impact statement}
This work enables training large language models entirely with low-rank factorized weights, producing models that match dense performance at equal compute while using fewer parameters. The main benefit is efficiency: lower memory, inference, and energy costs can reduce the financial and environmental burden of deploying foundation models and broaden access for those with limited hardware. As with any advance in efficient model training, these gains are dual-use, since cheaper training and inference also lower the barrier to misuse. Our contribution is methodological and introduces no capabilities beyond those of comparable dense models, so the associated risks are inherited from large language models in general rather than created here. We believe the efficiency and accessibility benefits outweigh these marginal risks.

\nocite{langley00}

\bibliography{example_paper}
\bibliographystyle{icml2026}

\newpage
\appendix
\onecolumn

\section{Algorithms}

We present two auxiliary algorithms utilized in our method. Algorithm~\ref{alg:newton_schulz} describes the Newton-Schulz iteration~\citep{jordan2024muon}, an efficient procedure for matrix orthogonalization. Algorithm~\ref{alg:power_iter} presents the power iteration method~\citep{vogels2019powersgd}, which we use to estimate spectral norms of the low-rank factors.

\begin{algorithm}
\caption{Newton-Schulz Orthogonalization}
\label{alg:newton_schulz}
\begin{algorithmic}[1]
\Require Gradient/Momentum matrix $G \in \mathbb{R}^{m \times n}$, iteration steps $k_{ns}=5$, numerical stability constant $\epsilon=10^{-7}$
\Require Newton-Schulz coefficients: $(a, b, c) = (3.4445, -4.7750, 2.0315)$
\Ensure Orthogonalized matrix $X_{k_{ns}} \in \mathbb{R}^{m \times n}$
\State $X_0 \gets G$ \Comment{Initialize with input matrix}
\State $s \gets \frac{1}{\|X_0\|_F + \epsilon}$ \Comment{Frobenius normalization factor}
\State $X_1 \gets sX_0$ \Comment{Normalized initialization}
\If{$m > n$} \Comment{Transpose for computational efficiency}
    \State $X_1 \gets X_1^\top$
\EndIf
\For{$i = 1$ \textbf{to} $k_{ns}$} 
    \State $A_i \gets X_i^\top X_i$ \Comment{Gram matrix computation}
    \State $B_i \gets bA_i + cA_i^2$ 
    \State $X_{i+1} \gets aX_i + X_iB_i$ \Comment{Newton-Schulz update step}
\EndFor
\If{$m > n$} \Comment{Restore original orientation}
    \State $X_{k_{ns}+1} \gets X_{k_{ns}+1}^\top$
\EndIf
\State \Return $X_{k_{ns}+1}$

\end{algorithmic}
\end{algorithm}

\begin{algorithm}[!]
\caption{PowerIter: Approximate Largest Singular Value and Left Singular Vector}
\label{alg:power_iter}
\begin{algorithmic}[1]
\Require Matrix $W \in \mathbb{R}^{p \times q}$, initial vector $u \in \mathbb{R}^{p}$ (normalized), iterations $k$ (default: 1)
\State $u \leftarrow u / \|u\|_2$ \Comment{Ensure initial normalization}
\For{$i = 1, \dots, k$}
    \State $v \leftarrow W^\top u$
    \State $v \leftarrow v / \|v\|_2$ \Comment{Right vector (normalized)}
    \State $u \leftarrow W v$
    \State $u \leftarrow u / \|u\|_2$ \Comment{Left vector (normalized)}
\EndFor
\State $\sigma \leftarrow u^\top W v$ \Comment{Rayleigh quotient approximation of $\sigma_{\max}$}
\State \Return $\sigma$, $u$
\end{algorithmic}
\end{algorithm}

\clearpage
\section{Ablations}
\label{sec:ablations}

We systematically evaluate the contribution of each component in our method using a Factorized Llama-94M model.

\subsection{Effect of Orthogonalization and Spectral Renormalization}
\begin{figure}[htbp]
    \centering
    \includegraphics[width=0.6\linewidth]{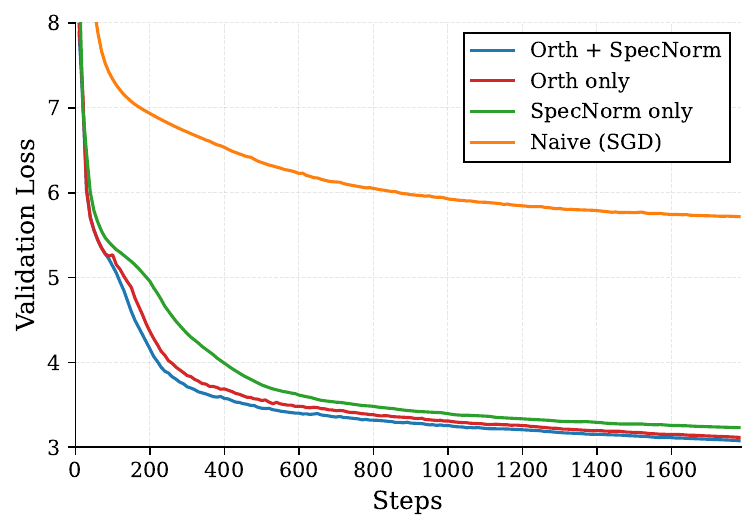}
    \caption{\textbf{Ablation Study of Orthogonalization and Spectral Renormalization Components.} Validation loss curves over 1,800 training steps for a Factorized Llama-94M model trained with different combinations of our method's core components: combined orthogonalization and spectral renormalization (\textcolor{blue}{blue}), orthogonalization only (\textcolor{red}{red}), spectral renormalization only (\textcolor{green}{green}), and naive SGD baseline (\textcolor{orange}{orange}). While both individual components substantially improve upon the baseline, their combination (\textcolor{blue}{blue}) achieves the lowest final validation loss (3.00), demonstrating that gradient orthogonalization and spectral renormalization provide complementary effects essential for stable low-rank training. The naive baseline exhibits severely impaired convergence, plateauing at ~7.0 validation loss, highlighting the critical importance of spectral control in factorized architectures.}
    \label{fig:abl-component_analysis}
\end{figure}

\begin{table}[htbp]
\centering
\begin{tabular}{@{}cccc@{}}
\toprule
Gradient Orthogonalization & Spectral Renormalization & Perplexity & Val loss \\ 
\midrule
$\times$ & $\times$ & 1042.02 & 6.95 \\
$\times$ & \checkmark & 24.01   & 3.18 \\
\checkmark & $\times$ & 20.95   & 3.04 \\
\checkmark & \checkmark & \textbf{20.26} & \textbf{3.00} \\ 
\bottomrule
\end{tabular}
\caption{\textbf{Ablation Analysis of Gradient Orthogonalization and Spectral Renormalization Components.} Systematic evaluation of each component's contribution to final model performance on Factorized Llama-94M. Gradient orthogonalization (Orth) ensures updates lie in the constraint radius, while spectral renormalization (SpecNorm) controls the constraint radius based on the spectral norms of the factorized weight matrices. The naive baseline (neither component) achieves 1042 perplexity and 6.95 validation loss. Activating spectral renormalization alone reduces val loss to 3.18 (54\% improvement), while orthogonalization alone achieves 3.04 (56\% improvement). The full method combining both components yields optimal performance at 3.00 validation loss, demonstrating that these mechanisms provide synergistic control.}
\label{tab:ablation-orth-specnorm}
\end{table}

Figure~\ref{fig:abl-component_analysis} and Table~\ref{tab:ablation-orth-specnorm} present an analysis of our two core components. The naive baseline fails catastrophically, while both components independently recover stable training. Combining orthogonalization and spectral renormalization achieves the best performance, confirming their complementary roles. Orthogonalization only benchmark yields the Muon~\cite{jordan2024muon} optimizer. 

\subsection{Effect of Rank Ratio}
\begin{figure}[htbp]
    \centering
    \includegraphics[width=0.6\linewidth]{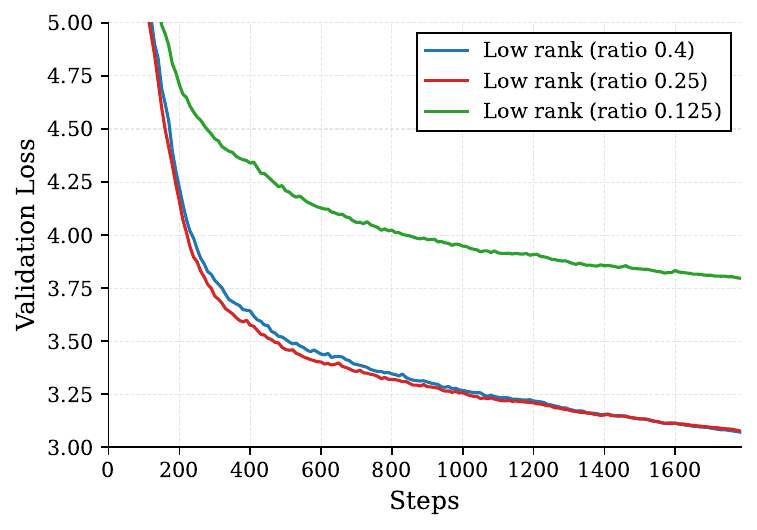}
    \caption{\textbf{Effect of Rank Ratio in Low-Rank Factorized Training.} Validation loss curves across 1,800 optimization steps for Factorized Llama-94M trained with different rank ratios under our method. Models with rank ratios 0.4 (\textcolor{blue}{blue}) and 0.25 (\textcolor{red}{red} ) exhibit nearly identical convergence dynamics with 0.4 being slightly better. In contrast, aggressive compression to rank ratio 0.125 (\textcolor{green}{green} ) results in fundamentally degraded learning dynamics. We choose 0.25 for all our experiments as a best trade off.}
    \label{fig:rank_ratio_ablation_plot}
\end{figure}

\begin{table}[b]
\centering
\begin{tabular}{@{}ccc@{}}
\toprule
\multicolumn{1}{l}{Rank Ratio} & \multicolumn{1}{l}{Perplexity} & \multicolumn{1}{l}{Val loss} \\ \midrule
0.25                           & 20.26                       & 3.00                     \\
0.4                            & 20.00                        & 2.99                     \\
0.125                          & 42.65                      & 3.75                    \\
\bottomrule
\end{tabular}
\caption{\textbf{Sensitivity of Low-Rank Training Performance to Rank Ratio Selection.} Validation loss and perplexity as a function of rank ratio (proportion of retained dimensions relative to full-rank parameterization) for Factorized Llama-94M with our method. The rank ratio of 0.4 achieves optimal performance, marginally outperforming 0.25. However, aggressive compression to 0.125 results in substantial degradation. We choose a rank ratio of 0.25 for all our experiments to balance parameter reduction and performance tradeoff.}
\label{tab:rank_ratio_ablation}
\end{table}

\begin{table}[htbp]
\centering
\begin{tabular}{@{}lccc@{}}
\toprule
Method            & $r = 0.25$ & $r = 0.4$ & $r = 0.5$ \\ \midrule
Dense (Full-Rank) & 3.070      & 3.070     & 3.070     \\ \midrule
Vanilla AdamW     & 3.274      & 3.196     & 3.171     \\
Self-Guided       & 3.240      & 3.207     & 3.187     \\
Spectron (Ours)   & \textbf{3.084} & \textbf{3.028} & \textbf{3.013} \\
\bottomrule
\end{tabular}
\caption{\textbf{Validation Loss Across Rank Ratios.} Validation loss for Factorized Llama-94M trained with different low-rank optimization methods at rank ratios $r \in \{0.25, 0.4, 0.5\}$. Dense full-rank training (3.070) is shown as a reference. Spectron consistently achieves the lowest loss among all low-rank methods, and surpasses the dense baseline at $r = 0.4$ and $r = 0.5$.}
\label{tab:rank_ratio_method_comparison}
\end{table}

\begin{figure}[t]
    \centering
    \includegraphics[width=0.8\textwidth]{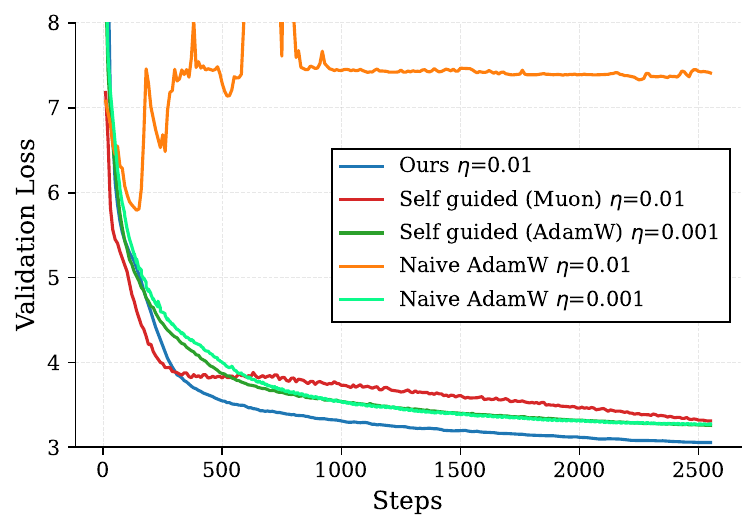}
    \caption{\textbf{Higher Learning Rates Destabilize Factorized Training in Naive training.} 
    Validation loss curves comparing different optimization methods for low-rank Factorized Llama-94M across learning rates $\eta \in \{0.001, 0.01\}$. 
    Naive AdamW (\textcolor{orange}{orange} ) diverges catastrophically at $\eta=0.01$ and converges slowly at $\eta=0.001$, demonstrating that standard optimizers cannot support aggressive learning rates for factorized models.
    Self-guided training (with AdamW~\cite{kingma2015adam}) improves stability through dense matrix guidance but still requires 25\% additional FLOPs and memory for auxiliary weights. We see little improvement on Self-guided training when using Muon~\cite{jordan2024muon} optimizer.  
    Our method (\textcolor{blue}{blue}) achieves stable, fast convergence at $\eta=0.01$ with only 1\% FLOPs overhead via spectral renormalization and orthogonalization, eliminating the training instability inherent to factorized matrices.}
    \label{fig:ablation_learning_rates}
\end{figure}

Figure~\ref{fig:rank_ratio_ablation_plot} and Table~\ref{tab:rank_ratio_ablation} examine sensitivity to the rank ratio. We defined rank ratio as the multiple used to set the low rank $r$ using the input dimension $n$ of a matrix sized $m \times n$. Rank ratios of $0.25 \times n$ and $0.4 \times n$ achieve comparable performance with $0.4$ being slightly better. Aggressive compression to $0.125 \times n$ substantially degrades learning. We adopt $0.25$ for all experiments to balance parameter efficiency with performance.

\begin{figure}
    \centering
    \includegraphics[width=0.8\linewidth]{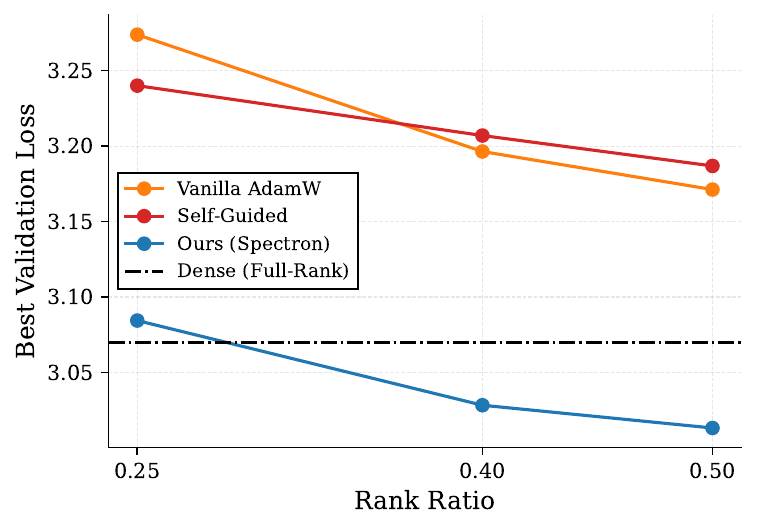}
    \caption{\textbf{Best Validation Loss Across Rank Ratios} Best validation loss for Factorized Llama-94M trained at rank ratios $r \in \{0.25, 0.4, 0.5\}$ under three low-rank optimization methods—Vanilla AdamW (\textcolor{orange}{orange}), Self-Guided (\textcolor{red}{red}), and our method Spectron (\textcolor{blue}{blue})—with dense full-rank training shown as a reference (\textbf{black} dash-dot line). Spectron consistently achieves substantially lower loss than both low-rank baselines at every rank ratio, and crosses below the dense full-rank reference at $r \approx 0.26$, surpassing it for all higher ranks.}
    \label{fig:rank_trend}
\end{figure}

In addition, Table~\ref{tab:rank_ratio_method_comparison} reports validation loss across optimization methods and rank ratios. We make two key observations: (a) Spectron consistently outperforms all low-rank baselines across every rank ratio, and (b) Spectron surpasses dense full-rank training at higher ranks. Performance continues to improve with rank as show in Figure~\ref{fig:rank_trend}.

\subsection{Effect of Learning Rate on Training Stability}
\label{apdx:ablation_stability}

Training low-rank factorized models presents a critical challenge: extreme sensitivity to learning rate selection.
Figure~\ref{fig:ablation_learning_rates} demonstrates how this instability manifests across different optimization approaches and learning rates. This behavior aligns with our analysis in Section~\ref{sec:spectral_renormalization}, where we show that large spectral norm fluctuations cause proportional changes in activation RMS, destabilizing training.

We observe that naive AdamW training fails completely at $\eta=0.01$, producing loss spikes that lead to divergence. Large activation changes during gradient updates cause this failure. While reducing the learning rate to $\eta=0.001$ achieves convergence, training proceeds slowly and inefficiently.

Self-guided training~\cite{wei2024building} exhibits mixed results depending on the optimizer choice. When combined with Muon~\cite{jordan2024muon}—an optimizer that applies gradient orthogonalization—self-guided training supports higher learning rates but produces inferior final performance. Self-guided training with AdamW~\cite{kingma2015adam} at lower learning rates achieves better results, motivating our choice of AdamW~\cite{kingma2015adam} for self-guided baselines in the main paper.

Our method combines orthogonalization with spectral renormalization to achieve both stable and fast convergence. At $\eta=0.01$, our approach converges smoothly without the instabilities that plague naive factorized training, while maintaining computational efficiency with only 1\% FLOPs overhead. This demonstrates that explicit spectral control enables factorized models to leverage aggressive learning rates that would otherwise cause training failure.

\subsection{Effect of factorizing fully connected layers only }

While our main experiments apply low-rank factorization to all non-embedding matrices, we conduct an ablation study to evaluate Spectron's effectiveness when restricting factorization exclusively to feedforward network (FFN) layers—the configuration adopted by \citet{wei2024building}. This controlled comparison directly addresses whether Spectron's advantages persist under the more constrained setting proposed by prior work.

\begin{figure}[t]
    \centering
    \includegraphics[width=0.9\linewidth]{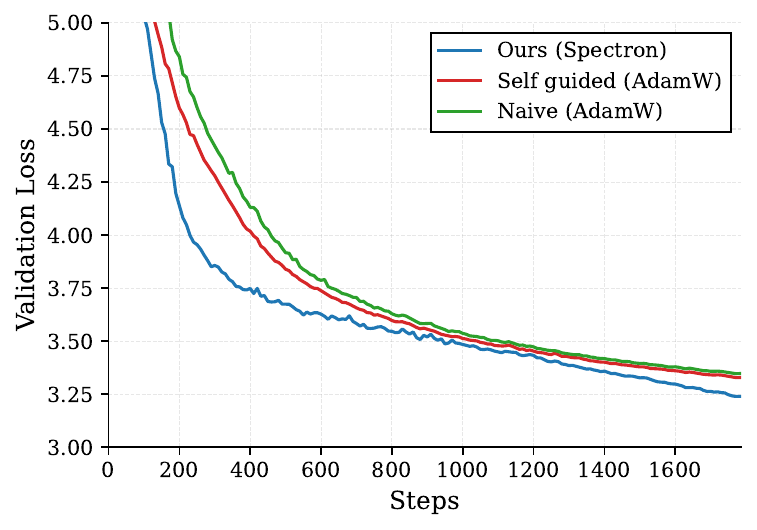}
    \caption{\textbf{Spectron Outperforms Baselines when only fully connected layers factorized to low rank} Validation loss curves comparing Spectron (\textcolor{blue}{blue}) against self-guided learning (\textcolor{red}{red} ) and naive AdamW training (\textcolor{green}{green} ) when applying low-rank factorization exclusively to feedforward layers. Spectron achieves lower validation loss throughout training and converges faster, surpassing self-guided learning  even within the setting proposed by~\citet{wei2024building}}
    \label{fig:ffn_only}
\end{figure}

Figure~\ref{fig:ffn_only} demonstrates that Spectron consistently outperforms both self-guided learning and naive AdamW training even when limiting low-rank factorization to FFN layers only. Spectron achieves a superior convergence properties within the exact architectural constraints proposed by \citet{wei2024building}. This result validates that Spectron's optimization strategy provides tangible benefits beyond simply expanding the scope of factorization, and confirms its effectiveness across different factorization schemes.

\section{Self-Guided Training }
\label{apn:self-guided}
The self-guided training technique addresses optimization challenges in low-rank factorized weight matrices by introducing an auxiliary dense parameterization that facilitates stable training dynamics~\cite{wei2024building}. 

The method computes the layer output as a weighted combination:
\begin{equation}
\mathbf{o} = \alpha \cdot W\mathbf{x} + (1-\alpha) \cdot A(B\mathbf{x}),
\end{equation}
where $\alpha$ follows a cosine decay schedule from 1 to 0 throughout training.

Initially, the dense matrix $W$ dominates the computation, providing stable gradient signals that circumvent pathological landscapes inherent to factorized structures. As $\alpha$ decreases, the model progressively transitions to the factorized representation $(A,B)$, which inherits feature specialization learned by $W$ through backpropagated gradients.

The dense matrix is initialized as
\begin{equation}
W_0 = A_0 B_0,
\end{equation}
ensuring no behavioral change at the start and enabling application at any training stage, including fine-tuning.

To reduce computational overhead, a stochastic variant employs probabilistic sampling:
\begin{equation}
\mathbf{o} =
\begin{cases}
\alpha \cdot W\mathbf{x} + (1-\alpha) \cdot A(B\mathbf{x}) & \text{with probability } p < \alpha, \\
A(B\mathbf{x}) & \text{otherwise},
\end{cases}
\end{equation}
where $p \sim \text{Uniform}(0,1)$ is sampled independently each forward pass.

This formulation reduces expected computational cost to approximately half the deterministic version while maintaining comparable accuracy. Applying this technique during half of training is said to increase total floating-point operations by 25\% relative to using solely the factorized parameterization ~\cite{wei2024building}.

\section{Scaling Laws: Parametric Function Approach}
\label{apdx:approach_3}

We complement our IsoFLOP analysis (Section~\ref{sec:scaling_laws}) with the parametric approach of \citet{hoffmann2022an}. This method directly models validation loss as a function of model size $N$ and training tokens $D$ using the classical risk decomposition:
\begin{equation}
    L(N,D) = E + \frac{A}{N^{\alpha}} + \frac{B}{D^{\beta}},
\end{equation}
where $E$ represents irreducible loss, and the power-law terms capture underfitting due to limited model capacity and limited data, respectively. We refer to Appendix D.2 of \citet{hoffmann2022an} for further details of this functional form. 

\paragraph{Fitting Procedure.} We fit this functional form to all 39 data points from our IsoFLOP experiments, spanning model sizes from 47M to 1.5B parameters. We minimize the Huber loss~\citep{huber1992robust} with $\delta=10^{-3}$ between predicted and observed log loss using scipy~\cite{2020SciPy-NMeth} L-BFGS-B~\cite{nocedal1980updating} algorithm, obtaining the following estimates:
\begin{align}
    A &= 1000, & \alpha &= 0.398, \\
    B &= 1000, & \beta &= 0.332, \\
    E &= 1.777. & &
\end{align}

\paragraph{Compute-Optimal Allocation.} Following \citet{hoffmann2022an}, we derive the compute-optimal model size and token count by minimizing $L(N,D)$ subject to the constraint $C = 6ND$. Here $C$ denotes the total compute budget. This yields:
\begin{equation}
    N_{\text{opt}}(C) \propto C^{\frac{\beta}{\alpha+\beta}} = C^{0.45}, \qquad D_{\text{opt}}(C) \propto C^{\frac{\alpha}{\alpha+\beta}} = C^{0.55}.
\end{equation}
These exponents closely match the Chinchilla scaling laws ($N_{\text{opt}} \propto C^{0.46}$, $D_{\text{opt}} \propto C^{0.54}$), but are consistently slightly smaller for optimal parameter size and larger for optimal token count. This indicates that, for a fixed compute budget, compute-optimal low-rank pretraining favors smaller models trained on more tokens compared to dense training. The consistency across two independent fitting approaches (IsoFLOP in Section~\ref{sec:scaling_laws} and parametric) strengthens confidence in this scaling characterization.

\section{Implementation details}
\label{apdx:implementation_details}

We describe our experimental setup and implementation choices in this section. All models follow the Llama-3 architecture with RoPE~\citep{su2024roformer} positional embeddings.  We use the LLama-2 (7B) tokenizer with a vocabulary size of 32,000. 

\subsection{Model Configurations}

For our main experiments, we evaluate three model scales: Llama-134M, Llama-500M, and Llama-780M. For low-rank variants, we apply a rank ratio of 0.25, reducing the effective rank to $0.25 \times n$ where $n$ is the input dimension. Table~\ref{tab:main_models} summarizes the architectural configurations for these models. 

\begin{table}[h]
\centering
\caption{Main model configurations used in our experiments.}
\label{tab:main_models}
\begin{tabular}{lcccc}
\toprule
Model & Hidden Size & Layers & Heads & Parameters \\
\midrule
Llama-134M & 768 & 12 & 12 & 134M \\
Llama-500M & 1280 & 20 & 20 & 500M \\
Llama-780M & 1536 & 24 & 24 & 780M \\
\bottomrule
\end{tabular}
\end{table}

\subsection{Scaling Law Experiments}

For scaling law analysis, we use base models ranging from 60M to 2.7B parameters. We apply the same rank ratio of 0.25 to create low-rank variants, yielding reduced models from 47M to 1.5B parameters. Table~\ref{tab:scaling_models} provides the complete architectural details for all model scales.

\begin{table}[h]
\centering
\caption{Model configurations for scaling law experiments. All models use the same number of key-value heads as query heads.}
\label{tab:scaling_models}
\small
\begin{tabular}{lcccc}
\toprule
Base Params & Hidden Size & Layers & Heads & Low-Rank Params \\
\midrule
60M & 512 & 8 & 8 & 47M \\
92M & 640 & 10 & 10 & 68M \\
134M & 768 & 12 & 12 & 94M \\
150M & 768 & 14 & 12 & 101M \\
220M & 896 & 16 & 14 & 143M \\
325M & 1024 & 20 & 16 & 200M \\
500M & 1280 & 20 & 20 & 297M \\
780M & 1536 & 24 & 24 & 454M \\
835M & 1536 & 26 & 24 & 484M \\
1.4B & 2048 & 24 & 16 & 780M \\
1.7B & 2048 & 30 & 16 & 943M \\
2.7B & 2560 & 32 & 20 & 1.5B \\
\bottomrule
\end{tabular}
\end{table}

\subsection{Training Setup}

We use cosine~\cite{loshchilov2016sgdr} scheduler with cycle length set to total training steps and first 5\% steps as warmup. We decay the learning rate to 0. We conduct hyperparameter sweeps across learning rates and weight decay values to ensure optimal performance. Learning rates are swept across three logarithmic scales ($10^{-1}, 10^{-2}, 10^{-3}$) and three linear multipliers (1$\times$, 5$\times$, 7$\times$ of each log scale) for all configurations. Weight decay~\cite{loshchilov2018decoupled} is swept from $10^{-1}$ to $10^{-3}$. We report the best results obtained from these sweeps for all experiments. When AdamW~\cite{kingma2015adam} is used we use $\beta_1$ and $\beta_2$ as 0.9 and 0.95. For Muon~\cite{jordan2024muon} experiments we use 0.95 for momentum. 

Following \citet{wei2024building}, we initialize low-rank models using spectral initialization~\cite{khodakinitialization} while applying standard decoupled weight decay regularization~\cite{loshchilov2018decoupled}. All models are trained on the Fineweb~\cite{penedo2024fineweb} corpus with a batch size of 512 sequences and a sequence length of 2048 tokens, yielding $512 \times 2048 = 1{,}048{,}576$ tokens per training step. We train the dense models Llama-134M, Llama-500M, and Llama-780M for 1786, 5664, and 8657 steps, respectively. For the factorized models, Factorized Llama-94M, Factorized Llama-297M, and Factorized Llama-454M, we train for 2556, 9537, and 14878 steps, respectively. All experiments are conducted on NVIDIA H100 GPUs with mixed precision using bf16.

\section{Wall-Clock Time Analysis}

\begin{table}[htbp]
\centering
\caption{Wall-clock training times (in seconds) for different low-rank optimization methods across model scales.}
\label{tab:wallclock-times}
\begin{tabular}{lrrr}
\toprule
\textbf{Method} & \textbf{Factorized Llama 94M (s)} & \textbf{Factorized Llama 297M (s)} & \textbf{Factorized Llama 454M (s)} \\
\midrule
Naive Low-rank (AdamW) & 1{,}965.82 & 17{,}355.04 & 36{,}677.68 \\
Self-Guided               & 2{,}369.59 & 18{,}525.23 & 39{,}067.47 \\
Spectron (Ours)           & 2{,}013.39 & 17{,}700.99 & 36{,}861.37 \\
\bottomrule
\end{tabular}
\end{table}

We compared this with multinode training on 2 nodes having 4 H100s each interconnected by infiniband. Spectron adds only $\sim$2--2.4\% wall-clock time over naive low-rank AdamW, confirming the Newton-Schulz iterations and power-method estimates are lightweight in practice and it is substantially less than Self-Guided ($\sim$6.5\%)~\citep{wei2024building}.

The wall-clock times for dense training are 1{,}473.61\,\textit{s} (Llama 134M), 15{,}233.55\,\textit{s} (Llama 500M), and 30{,}974.24\,\textit{s}(Llama 780M). The gap between dense and low-rank times is due to dense training using highly optimized kernels while our low-rank forward pass uses two unfused matmuls with launch overhead that outweighs FLOP savings at this scale; in addition, the FLOP-matched models train on more tokens and our data loading pipeline has not been specifically optimised. Custom fused kernels can close this gap.

\section{Comparison with recent Low rank training methods}

In this section we compare with recent low rank training methods \citet{mo2025parameter} and \citet{liu2025cola}.
We want to highlight an important distinction: CoLA modifies the model architecture by replacing weight matrices with autoencoder modules for low-rank activations, whereas Spectron and LoRO operate within the standard architecture. Despite this difference, we compare against both for completeness on a LLaMA-94M model.

\begin{table}[htbp]
    \centering
    \begin{tabular}{l c c r}
        \toprule
        \textbf{Method} & \textbf{Arch.\ Change} & \textbf{Perplexity} & \textbf{Wall Clock (s)} \\
        \midrule
        LoRO          & No  & 24.56 & 2{,}212 \\
        Spectron      & No  & 21.86 & 2{,}013 \\
        CoLA          & Yes & 24.21 & 2{,}479 \\
        CoLA + Spectron & Yes & 19.95 & 2{,}594 \\
        \bottomrule
    \end{tabular}
    \caption{Performance comparison of LoRO, Spectron, CoLA, and their combination on the LLaMA-94M model.}
    \label{tab:cola-comparison}
\end{table}

Spectron outperforms LoRO by 2.7 perplexity points while also being $\sim$10\% faster in wall-clock time. Furthermore, Spectron is complementary to architectural approaches like CoLA so combining the two yields the best result overall (19.95), demonstrating that our optimizer-level contribution is orthogonal to architecture-level low-rank methods and can be composed with them for additional gains.

\subsection{Extension of Spectron to Nonlinear Factorizations}

We show below how we extend Spectron to the nonlinear factorization used in CoLA~\citep{liu2025cola}.

 When a nonlinearity (here we consider ReLU~\citep{nair2010rectified}) is inserted between the factors, the effective mapping becomes
$W_{\text{eff}} = A \sigma(Bx).$
For ReLU (and element-wise gates like SiLU~\citep{elfwing2018sigmoid}in the approximate sense), this can be written as
$W_{\text{eff}} = A D B$
where $D = \operatorname{diag}(\mathbf{1}[Bx > 0])$ is a data-dependent binary diagonal mask. Since \(D\) is a binary diagonal matrix, $|D|_2 \leq 1$ always. The perturbation bound becomes
\begin{equation}
|\Delta W_{\text{eff}}|_2 \leq |\Delta A|_2 |D|_2 |B|_2 + |A|_2 |D|_2 |\Delta B|_2 + |\Delta A|_2 |D|_2 |\Delta B|_2.
\end{equation}
Using $|D|_2 \leq 1$, this reduces to exactly the same bound as the linear case shown in the main paper:
\begin{equation}
|\Delta W_{\text{eff}}|_2 \leq |\Delta A|_2 |B|_2 + |A|_2 |\Delta B|_2 + |\Delta A|_2 |\Delta B|_2.
\end{equation}

The activation mask drops out entirely. Consequently, the spectral LR scaling rule $\eta / (|A|_2 + |B|_2 + 1)$ derived for the linear factorization applies to CoLA's nonlinear factorization with the same structural guarantees if the non-linearity is ReLU. This achieves the same goal of controlling the spectral norm growth of the product $W$ without destabilizing the activations after applying $W$ to $x$.

\subsection{Comparison with ReLoRA and SST}

 In this section we compare our method with full-rank training methods which use low rank updates. We note that this has some important differences in original setup of our paper. We train a low-rank model from scratch without any auxiliary weights, whereas ReLoRA \citep{lialin2024relora} requires a full-rank warm start and periodical merges using auxiliary weights, and SST \citep{zhao2025sparse} uses a spectral parameterization that decouples singular values from singular vectors and keeps the $U \Sigma V$ parameterization in memory.

We ran ReLoRA~\cite{lialin2024relora} and SST~\citep{zhao2025sparse} both with and without Spectron. Our experiments show that Spectron's normalization provides complementary benefits even when combined with ReLoRA and SST.

\begin{table}[htbp]
\scriptsize
\centering
\caption{Perplexity comparison of Spectron against ReLoRA~\citep{lialin2024relora} and SST~\citep{zhao2025sparse} on a 134M base model (94M trainable parameters).}
\label{tab:relora-sst-comparison}
\begin{tabular}{lcccccc}
\toprule
\textbf{Method} & \textbf{Warm start} & \textbf{Base params} & \textbf{Total params (in memory)} & \textbf{Trainable params} & \textbf{Perplexity} & \textbf{Reduction (\%)} \\
\midrule
Spectron (Our setup) & No  & 134M & 94M     & 94M & 21.86 & -- \\
ReLoRA               & Yes & 134M & 162.14M & 94M & 26.06 & -- \\
ReLoRA + Spectron    & Yes & 134M & 162.14M & 94M & 22.70 & 12.90 \\
SST                  & No  & 134M & 175.54M & 94M & 29.40 & -- \\
SST + Spectron       & No  & 134M & 175.54M & 94M & 23.40 & 20.40 \\
\bottomrule
\end{tabular}
\end{table}

The results show that Spectron delivers a 12.9\% perplexity reduction on ReLoRA and a 20.4\% reduction on SST, demonstrating that our spectral-norm stabilization is complementary to both dynamic-rank and spectral-parameterization approaches.  These auxiliary experiments strengthen the paper by situating Spectron's contribution relative to alternative training strategies.

\section{Spectral norm variation across layers}
\begin{figure*}[!ht]
    \centering
    \includegraphics[width=0.7\linewidth]{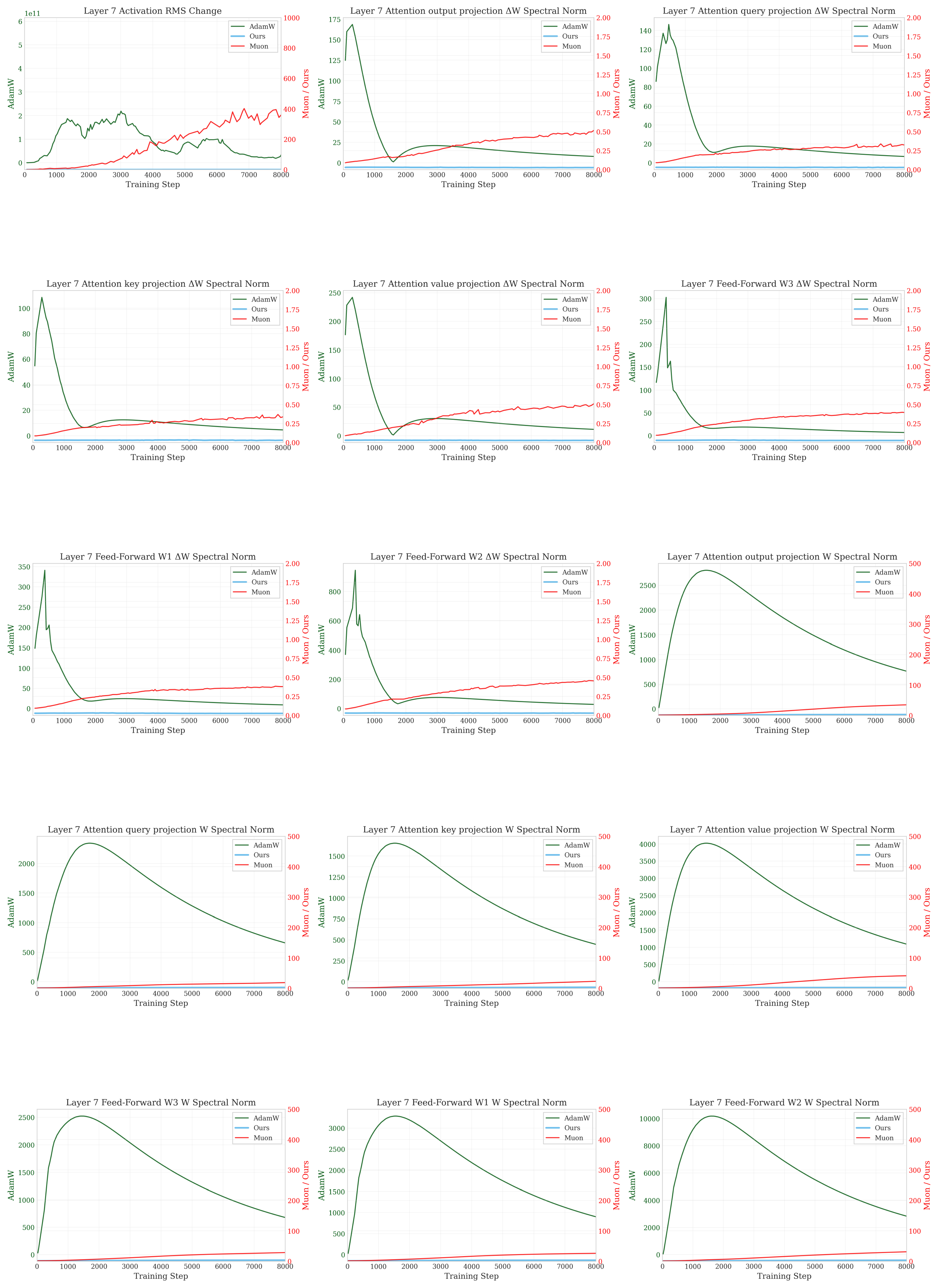}
    \caption{Spectral norm variation across layers}
    \label{fig:spec_rebuttal}
\end{figure*}

\clearpage
\section{Training curves}

\begin{figure}[!ht]
    \centering
    \begin{subfigure}[t]{0.9\linewidth}
        \centering
        \includegraphics[width=\linewidth]{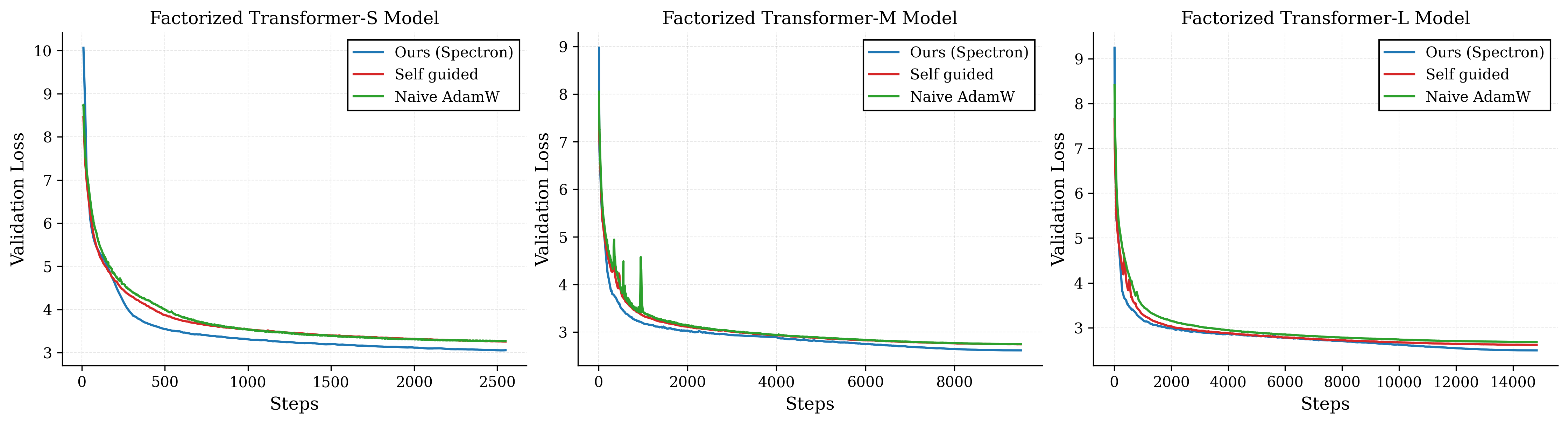}
        \caption{Training curves of our baseline comparisons in Section~\ref{sec:experiments}}
        \label{fig:curves-self-guided}
    \end{subfigure}
    \vspace{0.5em}   
    \begin{subfigure}[t]{0.9\linewidth}
        \centering
        \includegraphics[width=\linewidth]{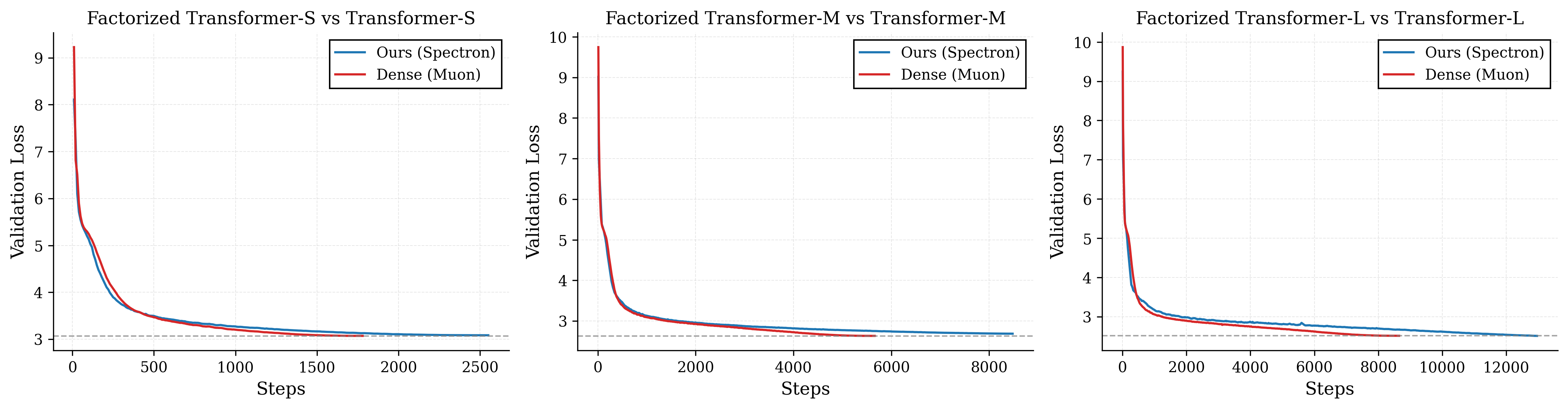}
        \caption{Training curves of our comparisons with dense models in Section~\ref{sec:experiments}}
        \label{fig:curves-dense}
    \end{subfigure}
    \vspace{0.5em}
    \begin{subfigure}[t]{0.9\linewidth}
        \centering
        \includegraphics[width=\linewidth]{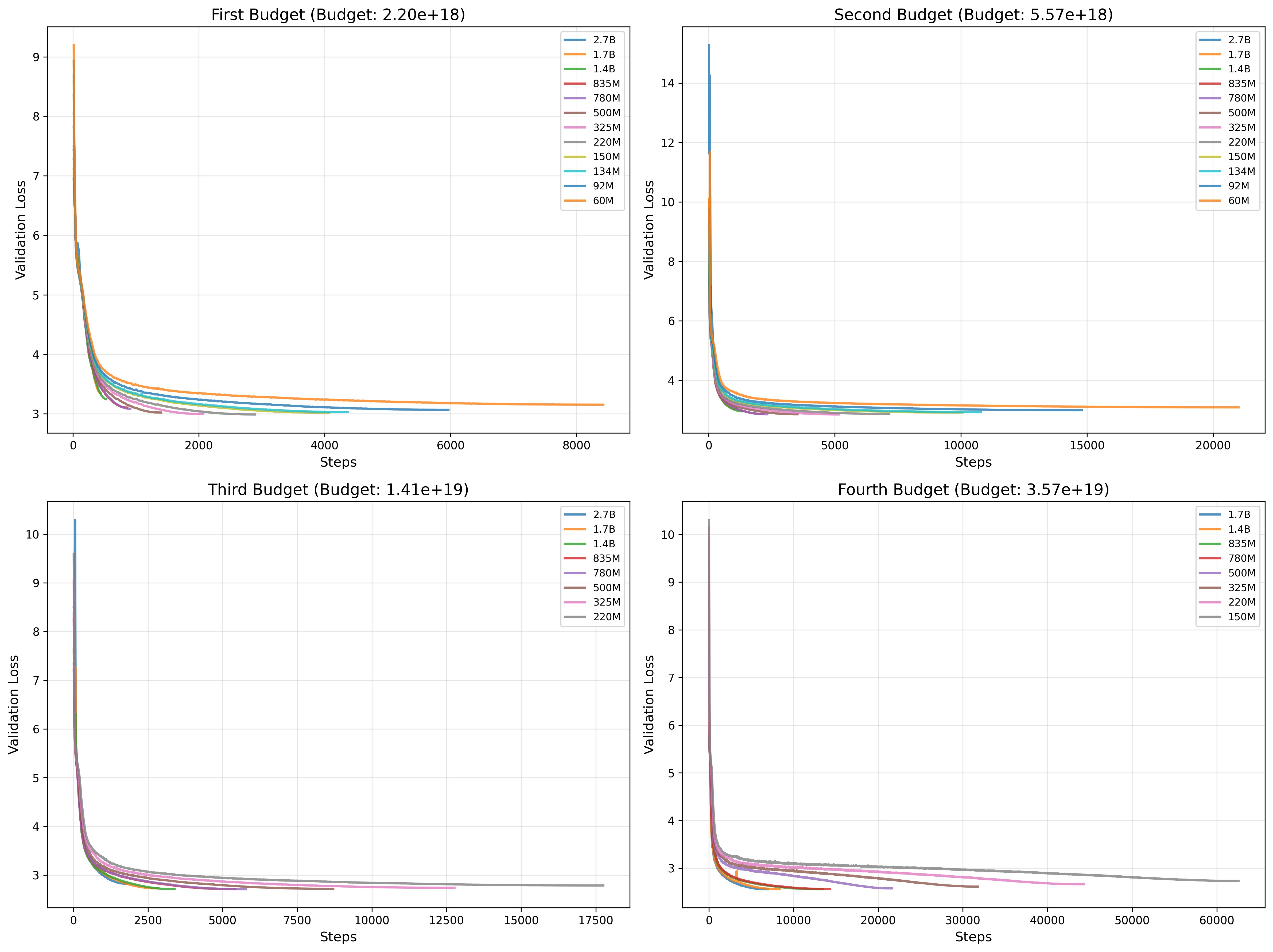}
        \caption{Training curves of our optimal model allocation experiments in Section~\ref{sec:scaling_laws}}
        \label{fig:curves-scaling}
    \end{subfigure}
    \caption{Training curves across different experimental settings}
    \label{fig:training-curves-all}
    \vspace{-3em}
\end{figure}


\end{document}